\documentclass{article}

    \PassOptionsToPackage{numbers, compress, sort}{natbib}

    \usepackage[preprint]{neurips_2025}

\usepackage[utf8]{inputenc} 
\usepackage[T1]{fontenc}    
\usepackage{hyperref}       
\usepackage{url}            
\usepackage{booktabs}       
\usepackage{amsfonts}       
\usepackage{nicefrac}       
\usepackage{microtype}      
\usepackage{xcolor}         

\usepackage{amsmath}
\usepackage{amsthm}
\usepackage[createShortEnv]{proof-at-the-end}
\usepackage{graphicx}
\usepackage{multirow}
\usepackage{makecell}
\usepackage{longtable}

\usepackage[capitalise, noabbrev]{cleveref}
\crefformat{footnote}{#2\footnotemark[#1]#3}

\usepackage{mathtools}
\usepackage{algorithm}
\usepackage{algorithmic}
\DeclarePairedDelimiter\abs{\lvert}{\rvert}%

\newtheorem{theorem}{Theorem}
\newtheorem*{theorem*}{Theorem}

\newtheorem{subclaim}{Claim}[theorem]
\crefname{subclaim}{Claim}{Claims} 
\newtheorem*{claim*}{Claim}

\title{Set-LLM: A Permutation-Invariant LLM}

\author{%
  Beni Egressy \\
  Heidelberg Institute for Theoretical Studies \\
  \texttt{egressbi@h-its.org} \\
  \And
  Jan Stühmer \\
  Heidelberg Institute for Theoretical Studies \\
  IAR, Karlsruhe Institute of Technology \\
  \texttt{jan.stuehmer@h-its.org} \\
}

\begin{document}

\maketitle

\begin{abstract}
While large language models (LLMs) demonstrate impressive capabilities across numerous applications, their robustness remains a critical concern. This paper is motivated by a specific vulnerability: the order sensitivity of LLMs. This vulnerability manifests itself 
as the order bias observed when LLMs decide between possible options (for example, a preference for the first option) and the tendency of LLMs to provide different answers when options are reordered. The use cases for this scenario extend beyond the classical case of multiple-choice question answering
to the use of LLMs as automated evaluators in AI pipelines, 
comparing output generated by different models. 
We introduce Set-LLM, a novel architectural adaptation for pretrained LLMs that enables the processing of mixed set-text inputs
with permutation invariance guarantees.
The adaptations involve a new attention mask and new positional encodings specifically designed for sets.
We provide a theoretical proof of invariance and demonstrate through experiments that Set-LLM can be trained effectively, achieving comparable or improved performance and maintaining the runtime of the original model, while eliminating 
order sensitivity.
\end{abstract}

\section{Introduction}



\begin{figure}[tbh]
    \centering
    \includegraphics[width=0.99\linewidth]{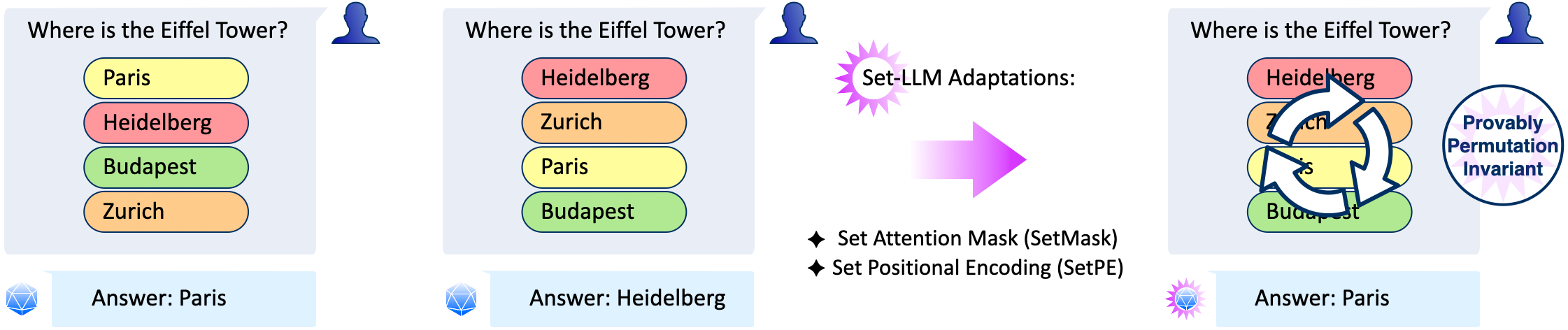}
    \caption{
    An example of the vulnerability of LLMs to choice permutations. The LLM's response 
    changes simply due to a reordering of the answer options. (Example for illustrative purposes only.) Set-LLM eliminates this vulnerability by building invariance directly into the model architecture.
    }
    \label{fig:intro_example}
\end{figure}

The remarkable achievements of Large Language Models (LLMs) in recent years \cite{grattafiori2024llama-short, achiam2023gpt, jiang2023mistral} have propelled their adoption across a wide range of applications, including safety-critical and sensitive domains such as medicine and finance \cite{zheng2025llminmedicinesurvey, li2023llminfinancesurvey}. 
%
%
%
As such, the eye-catching drops in performance from adversarial attacks can be all the more alarming 
\citep{shayegani2023-llm-vulnerabilities-survey, guo-etal-2021-gradient-adversarial-attacks}. One such attack, shown in \cref{fig:intro_example}, is as trivial as permuting the choices in multiple-choice questions, which \citet{zongfool} demonstrate can degrade an LLM's performance from ``good'' to worse than random.


This sensitivity to input order becomes even more critical given the increasing reliance on LLMs to compare and evaluate the output of other LLMs \citep{dubois2023alpacafarm-judge-order-bias, wang-etal-2024-large-language-models-fair-judge-order-bias, kim-etal-2025-biggen-llm-judge, zheng2023judgingLLM-as-a-Judge-MT-Arena}. 
Indeed, LLM-as-a-judge is widely used as an evaluation metric \citep{zheng2023judgingLLM-as-a-Judge-MT-Arena, badshah2024reference-LLMs-as-Judges-for-Free-Form-text}, and is also used
to annotate LLM-generated output to create new data sets and to decide between possible reasoning paths when solving complex problems \citep{gu2024survey-on-llm-as-a-judge, badshah2024reference-LLMs-as-Judges-for-Free-Form-text, gao2024strategyllm-LLM-judge-for-reasoning, liang-etal-2024-encouraging-LLM-judge-for-reasoning}. This inherent order sensitivity directly undermines the reliability of these pipelines.

We propose Set-LLM\footnote{All code is available under open licenses at <link added upon publication>.}, a permutation-invariant LLM architecture that eliminates this problem entirely. 
Set-LLM guarantees consistent responses by building permutation invariance directly into the model architecture.
Moreover, these guarantees are achieved without sacrificing model performance or increasing model complexity\footnote{Due to minor computational inconsistencies when changing the order of (mathematically order-independent) operations on hardware, and the accumulation of minor errors in deep neural networks, our models have to be run with a higher model precision to guarantee invariance. This adds a constant factor overhead to runtime costs.}, demonstrating that no trade-offs are necessary.

Set-LLM is based on the well-known observation that the attention mechanism underpinning all of the recent LLM architectures is permutation invariant by construction. In fact, to force them to take into account the order of the input tokens, almost all models use some form of positional encoding \citep{su2024roformer-ROPE, shaw-etal-2018-self-relative-PE, vaswani2017attention}.
However, \citet{kazemnejad2023impactNoPE} prove that even without positional encodings, the causal attention mask used in decoder-only LLMs is sufficient to completely reconstruct the input order.
We therefore remove positional encoding and causal masks as the first steps towards Set-LLM. 


Our complete approach comprises four steps: 
(1)~Removing sequential position encoding, 
(2)~Removing the causal mask, 
(3)~Adding permutation-invariant set position encoding (SetPE), and 
(4)~Adding a permutation-invariant set attention mask (SetMask). 
These steps are illustrated in \cref{fig:setMask_and_setPE}. Together, the first two steps result in a \emph{bag-of-words} (BoW) model, i.e., a model with no information about the token order, and the third and fourth steps add the order information we want back in.

We prove that Set-LLM is permutation-invariant, meaning it is guaranteed to give the exact same output whatever input order is chosen.
Moreover, we demonstrate that the Set-LLM adaptations can be combined with different decoder-only LLMs
and do not depend on specific architectures or model versions, which can quickly become outdated. 
We run experiments with five different base models on four multiple-choice datasets to show the approach's versatility in practice. 

We summarize the main contributions as follows:

\begin{itemize}
    \item We propose Set-LLM, the first permutation-invariant decoder-only LLM.
    \item We prove Set-LLM guarantees robustness to permutations, eliminating order sensitivity.
    \item We demonstrate that Set-LLM can be trained effectively with different base LLMs, consistently matching or outperforming the base models with random-order inputs, and significantly outperforming them with adversarial-order inputs, all without runtime overhead.
    \item This enables a more robust and efficient evaluation framework for multiple-choice question answering and for the use of LLMs as automated evaluators.
\end{itemize}

\section{Background: transformers and positional encoding}

The Set-LLM adaptations involve changes to the attention mask and positional encoding of transformer-based LLMs. We first describe these components before introducing Set-LLM.

\subsection{Attention scores}

Most state-of-the-art LLMs are based on the transformer architecture, made up of multiple attention layers stacked on top of one another \citep{vaswani2017attention}. Given the $d$-dimensional representation $X \in \mathbb{R}^{N \times d}$ of $N$ tokens, raw (unnormalized) attention scores (or weights) are calculated as
\begin{equation}
\label{eqn:attention}
    Z = \text{attn}(X,X, W_Q, W_K, d_K) = XW_{Q} (XW_{K})^T / \sqrt{d_K},
\end{equation}
where $W_{Q}, W_{K} \in \mathbb{R}^{d \times d_k}$ are the query and weight matrices, respectively, and $d_K$ is a scaling factor often chosen to be the dimension of the keys.

In a causal transformer, attention scores are masked to ensure that tokens can only \emph{attend} to preceding tokens, before being normalized through a softmax layer. 
Masking is usually represented by a matrix of $1$'s and $0$'s, $M \in \{0,1\}^{N \times N}$, where $M_{ij} = 1$ if token $i$ can \emph{attend} to token $j$.
However, masking can also be denoted by a directed graph $G^M=(V,E)$, where $(j,i) \in E(G^M) \iff M_{ij} = 1$, 
i.e., if information flows from token $j$ to token $i$. 
So for a causal mask, $(j,i) \in E(G^M)$ if and only if $j \leq i$.
Let $\mathcal{N}_i = \{j \mid (j,i) \in E(G^M)\}$ denote the \emph{neighborhood} (or \emph{field of view}) of the token $i$. 
Then we can write the normalized attention weights as
\begin{equation}
    A_{ij} = \text{softmax}_{G^M}(Z_{ij}) = \frac{\text{exp}(Z_{ij})}{\sum_{k \in \mathcal{N}_i} \text{exp}(Z_{ik})} = \frac{exp(Z_{ij})}{\sum_k M_{ik}exp(Z_{ik})}, 
\end{equation}
if $(j,i) \in E(G^M)$, and $A_{ij} = 0$ otherwise.

LLMs all use some kind of masking.
Decoder-only architectures use a causal attention mask, whereas encoder-decoder architectures, such as T5 \citep{raffel2020exploringT5}, used bidirectional (or fully-connected) attention for the prefix (or prompt) and causal attention for the output (or response).
We refer to this as \emph{prefix masking}. \cref{fig:setMask_and_setPE} illustrates the causal and prefix masks in both the matrix and graph forms.

Given normalized attention weights, the attention layer is completed by taking weighted averages of the token neighborhoods:
\begin{equation}
    X^{(t+1)} = A^{(t)} X^{(t)} W_V,
\end{equation}
where $W_V$ is the value matrix, and the superscript indicates the model layer.

\subsection{Positional encoding}

In addition to the causal mask, 
transformers use positional encoding to introduce positional information. There are two common variants: \emph{absolute positional encodings} and \emph{relative positional encodings}. 
Absolute positional encodings
assign consecutive integers to the tokens, starting at 0. These are usually embedded with an encoder layer and concatenated with the corresponding input token embeddings. 
The neural network can use these embeddings to ``understand'' word order.
The attention scores depend on the token embeddings $X_i, X_j$ and the absolute positions $i$ and $j$:
\begin{equation}
    Z_{ij} = \text{attn}_{abs}(X_i, X_j, W_Q, W_K, d_K, i, j).
\end{equation}
The positional information is often incorporated into the token embeddings, so row $i$ of $X$ is $x_i = \psi(q_i, i)$ for some function $\psi:\mathcal{T} \times \mathbb{N} \rightarrow \mathbb{R}^{d}$, and there is no further dependence on $i$, reducing $\text{attn}_{abs}$ to \cref{eqn:attention}.
We formulate our proofs using this notation.

On the other hand, relative positional encodings use the relative distance ($i-j$) of tokens 
in attention calculations. The attention scores depend on the token embeddings $X_i$, $X_j$, and the relative distance:
\begin{equation}
    Z_{ij} = \text{attn}_{rel}(X_i, X_j, W_Q, W_K, d_K, i-j).
\end{equation}
In this sense, relative position encodings can be seen as a special case of absolute position encodings, where a translation symmetry on positions is enforced, i.e. shifting the absolute positions by $m$ does not change the attention scores and therefore the attention layer outputs:
\begin{equation}
    \text{attn}(X_i, X_j, W_Q, W_K, d_K, i-j) = \text{attn}(X_i, X_j, W_Q, W_K, d_K, (i+m)-(j+m)).
\end{equation}

\section{Methods: Set-LLM}
\label{sec:methods}

\begin{figure}
    \centering
    \includegraphics[width=0.9\linewidth]{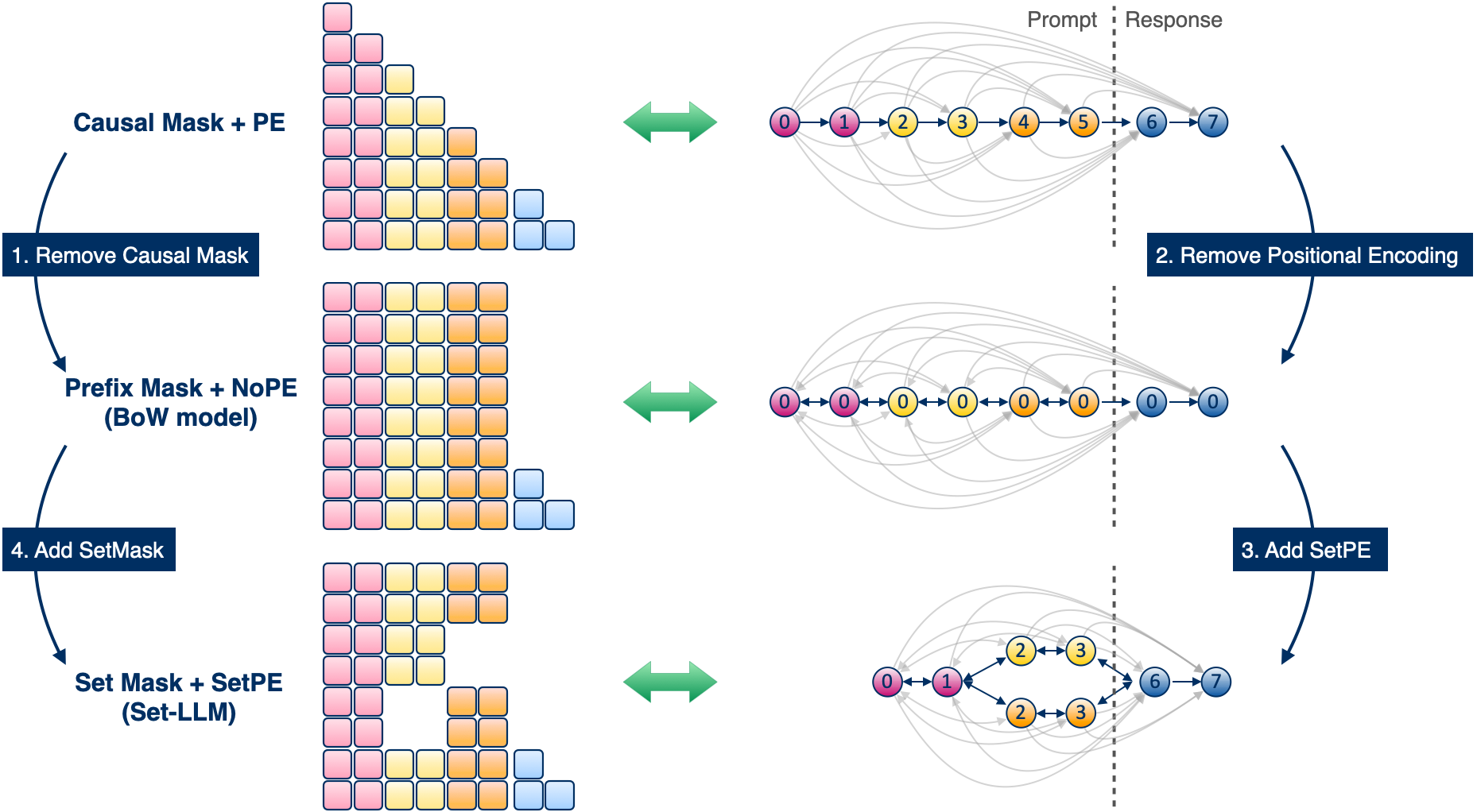}
    \vspace{-0.1cm}
    \caption{Three types of attention masks and their corresponding directed graphs. The colored squares on the left indicate attention scores that are not masked. 
    For example, in a causal mask, the $4^{\text{th}}$ token attends to the first $4$ tokens, and the remaining tokens are masked.
    The circles on the right represent the tokens as nodes of an attention graph. 
    Red, orange, and yellow tokens correspond to the prompt, and blue tokens correspond to the response. 
    Orange and yellow tokens corresponds to elements of a set.
    Causal masks are standard in decoder-only LMs, whereas prefix masks are used in bidirectional encoder-decoder LMs. 
    SetMask is introduced in this work.
    Some edges are grayed out and self-loops are omitted to improve clarity.
    In addition, the figure shows three types of token positions, standard consecutive positions (PE), \emph{no positional encoding} (NoPE), and \emph{set position encoding} (SetPE). These are indicated by the numbers inside the token nodes on the right.}
    \label{fig:setMask_and_setPE}
    \vspace{-0.1cm}
\end{figure}

LLMs are used for a variety of tasks, and many of them have sets in the input instructions. 
This includes answering multiple-choice questions and comparing LLM-generated outputs.
For example, $\left[ \{\text{``Which city is the capital of France:''}\}, \{ \text{``Budapest''}, \text{``Paris''}, \text{``Heidelberg''}, \text{``Zurich''} \} \right]$. 

More generally, if $\mathcal{T}$ denotes the token vocabulary, then 
a mixed set-text instruction can be written as $q =\left[ q_0, q_1, \ldots, q_n \right]$, where each $q_i$ is a set of token sequences:  
$q_i = \left\{ s_0, s_1, \ldots, s_{n_i} \right\}$ and $s_j = [\tau_0, \tau_1, \ldots, \tau_{n_{i,j}}]$, with all $\tau_i \in \mathcal{T}$. If $q = [q_0]$ and $\abs{q_0} = 1$, then we are back to the regular case where an instruction is a single sequence of tokens. In the above example, depending on the tokenization, we might have $q_0=\{\left[\text{``Which''}, \text{`` city''}, \ldots \right]\}$ and $q_1 = \{ \left[\text{``Buda''}, \text{``pest''}\right], \left[\text{``Paris''}\right], \ldots \}$.

For ease of notation, we use global indexing, where tokens in a set of choices are also numbered consecutively (in the default order in which the set is provided). 
Then the tokens of $q$ are $[t_0, t_1, \ldots, t_N]$, with $t_i \in \mathcal{T}$, where $N$ is the total number of tokens in $q$. We use $s(t_i)$ to denote the token sequence containing the token $t_i$ and $q(t_i)$ to denote the set containing $s(t_i)$.

Since LLMs take sequences as input, one would typically force an order onto sets within mixed 
input. However, ideally, we would like an LLM, whose output does not depend on this order.
Our proposed approach, Set-LLM, achieves this in four steps: 
(1) Remove sequential position encoding, that is, set all positions to $0$ (also called NoPE -- No Positional Encoding);
(2) Remove the causal mask and replace it with a prefix mask;
(3) Add permutation-invariant set position encoding (SetPE); and 
(4) Add permutation-invariant set attention masking (SetMask).
The steps are illustrated in \cref{fig:setMask_and_setPE}.

The first two steps already guarantee set permutation invariance. In fact, they create a bag-of-words (BoW) model that ignores the order of all input tokens, not just the order of elements within a set.
Clearly, BoW models have their limitations, since word order is critical to language.
Steps (3) and (4) are therefore crucial in reintroducing the order information within the different spans of text.

\begin{figure}
    \centering
    \includegraphics[width=0.9\linewidth]{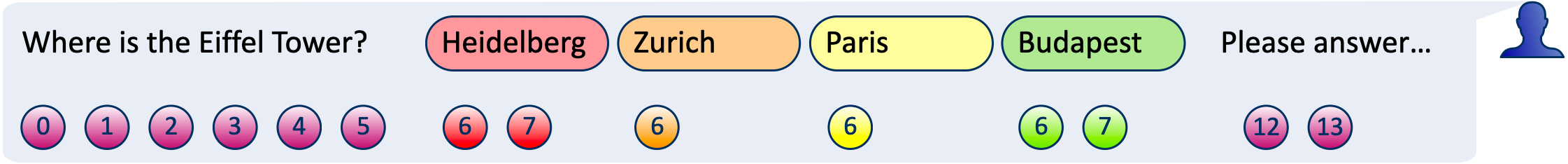}
    \vspace{-0.1cm}
    \caption{An example of a multiple-choice question with set positional encoding (SetPE) positions.}
    \label{fig:setPE_example}
    \vspace{-0.1cm}
\end{figure}

\subsection{Set position encoding (SetPE)}

While BoW models inherently disregard the order of set elements, they also disregard the word order within the text.
To overcome this limitation, we introduce set position encoding (SetPE). For SetPE, we generate \emph{SetPE positions} and use them to calculate absolute or relative positional encodings. The idea is to number tokens consecutively, but to number elements of a set from the same starting position.
An example is provided in \cref{fig:setPE_example}, along with pseudocode in 
\cref{app:setPE_alg}.

The example shows how SetPE positions align with standard absolute positions for ``regular'' text (starting at 0). However, when a set of options appears (e.g., at position 6), all options within that set are numbered consecutively from that starting position. 
This ensures that no order is forced upon the options, but the token order within the options is clear.
Positions resume with their absolute positions after a set of options (continuing with $12$ in the example). 
We denote SetPE positions by the function $\phi$ (or \texttt{pos} in the pseudocode), i.e., the position of the $i^{\text{th}}$ token of query $q$ is written $\left.\phi(q)\right|_{i} = \texttt{pos}[i]$.

Given the SetPE positions, we can calculate absolute or relative positional embeddings.
When using absolute positional encoding, the SetPE positions simply replace the absolute positions and are encoded and concatenated with the input token embeddings. 
When using relative positional encoding, the difference between SetPE positions is used to calculate relative positional embeddings rather than the difference between absolute positions. All LLMs in this paper specifically use RoPE \citep{su2024roformer-ROPE}. In this case, the SetPE positions determine the angles of rotation for the token embeddings. 

\subsection{Set attention mask (SetMask)}

\begin{figure}
    \centering
    \includegraphics[width=0.75\linewidth]{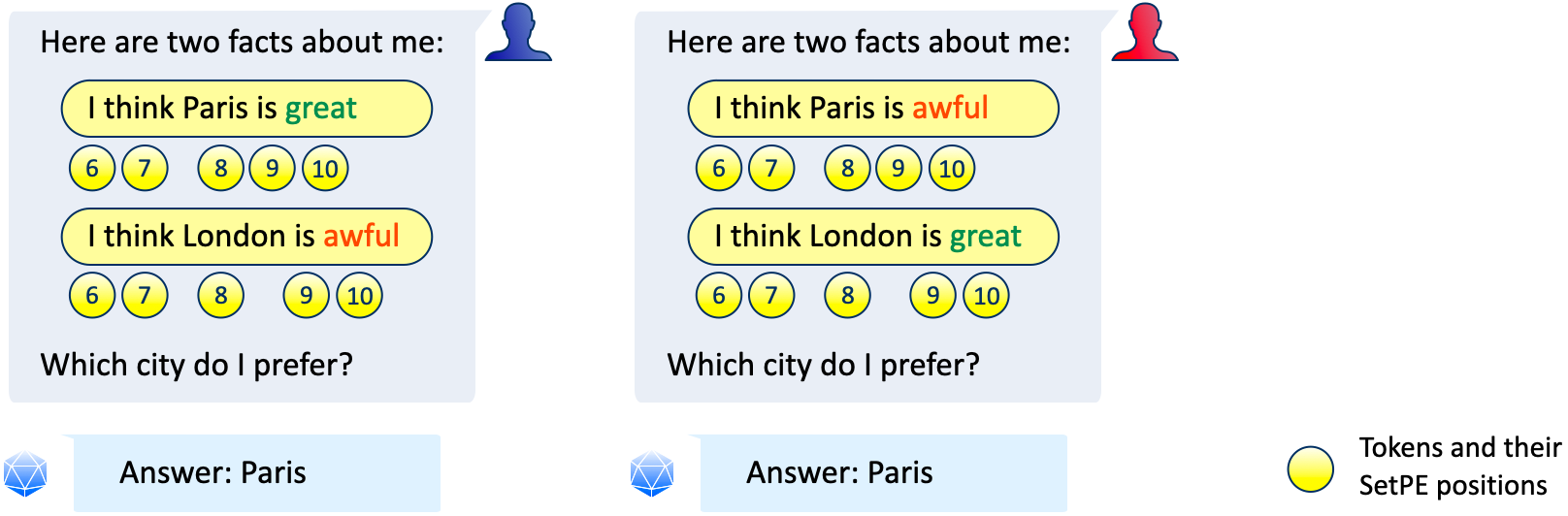}
    \caption{A failure case for an LLM with set position encoding (SetPE) but without set attention mask (SetMask). SetPE positions are shown for the tokens that are part of the set (of facts). 
    Without the SetMask, the model is unable to distinguish the two inputs.
    The model can not ``know'' which position $8$ token belongs to which position $10$ token.}
    \label{fig:setPE_fail}
\end{figure}

When using prefix mask, SetPE is insufficient for distinguishing mixed inputs. Consider the example of two inputs in \cref{fig:setPE_fail}, which contain two sets of opposing facts. Clearly, one person prefers Paris, while the other prefers London, but an LLM with prefix masking and SetPE will output the same next token distributions. 
To see this, note that all tokens in the two cases receive the same SetPE positions, although \emph{great} and \emph{awful} have been switched.
Since the prefix mask is fully connected, the attention layer outputs are then identical up to switching the two respective embeddings.

To address this shortcoming, we introduce set attention masking (SetMask). The idea is to use the attention mask to distinguish between the two cases in \cref{fig:setPE_fail}. SetMask is constructed by starting with a 
prefix mask
and removing all edges between tokens of different elements of the same set (that is, setting the respective matrix entries to $0$). An illustration in both matrix and graph forms can be seen in \cref{fig:setMask_and_setPE} (bottom). More precisely,
\begin{equation}
    M_{ij} = \begin{cases}
        0 \quad \text{if $q(t_i) = q(t_j)$ and $s(t_i) \neq s(t_j)$}\\
        1 \quad \text{else.}
    \end{cases}
\end{equation}
Since generation is autoregressive, response tokens can only attend to prompt tokens and preceding response tokens. SetMask is therefore extended to the response in the same way as causal mask and prefix mask.
%
With SetMask, the tokens corresponding to \emph{great} and \emph{awful} in \cref{fig:setPE_fail} have different neighborhoods in the two inputs, leading to different embeddings and different next token distributions.\looseness=-1

\subsection{Permutation invariance}

We claim that by construction, an attention layer with SetPE and SetMask is set permutation equivariant. 
Then, since all intermediate layers are set permutation equivariant and the final layer of an LLM is permutation invariant,
it follows that the whole network is set permutation invariant \citep{bronstein2021geometric}. 

\begin{theoremE}[Set Permutation Equivariance][end, restate, text link=]
\label{thm:permutation-invariance}
    Let $\pi$ be a permutation corresponding to permuting elements of sets in a mixed set-text input, and let $P$ be the corresponding permutation matrix. If $X^{(t+1)} = A^{(t)} X^{(t)} W_V$ is the output of an attention layer with SetPE, SetMask, and absolute positional encoding, then $\tilde{X}^{(t+1)} = P X^{(t+1)}$. In other words, attention with SetPE and SetMask is equivariant to set permutations of mixed set-text input.
\end{theoremE}

\begin{proofE}

\begin{subclaim}
\label{claim:X=PX}
    $\tilde{X} = PX$
\end{subclaim}
First note, that by the definition of SetPE, SetPE positions will be the same before and after the set permutation is applied to the input. This is because all token sequences remain in the same set after a set permutation, and all token sequences within a set are numbered consecutively from the same starting index. 
If $\ \tilde{} \ $ denotes variables after permuting the input, then
$\tilde{x}_{\pi(i)} = \psi(t_i, \texttt{pos}[i]) = x_i$. 
Tokens that are not part of a set are not moved by a set permutation, $\pi(i) = i$ for these tokens and again $\tilde{x}_{\pi(i)} = \psi(t_i, \texttt{pos}[i]) = x_i$.
Therefore, $\tilde{X} = PX$. \qed

\begin{subclaim}
\label{claim:Z=PZPT}
    $\tilde{Z} = P Z P^T$
\end{subclaim}
By \cref{claim:X=PX}, we have
\begin{align*}
    \tilde{Z} \quad
    &= attn(\tilde{X}, \tilde{X}, W_Q, W_K, d_K) \\
    &= attn(PX, PX, W_Q, W_K, d_K) \\
    &= PXW_Q (PXW_K)^T / \sqrt{d_K} \\
    &= PXW_Q W_K^TX^TP^T / \sqrt{d_K} \\
    &= P \left(XW_Q W_K^TX^T / \sqrt{d_K} \right)P^T 
    \quad= P Z P^T. 
\end{align*} 
\qed

\begin{subclaim}
\label{claim:M=PMPT}
    $\tilde{M} = PMP^T$
\end{subclaim}
Though this is not the case for general permutations of the tokens, we show that this is the case when $\pi$ is restricted to permuting elements within the sets $q_i$. 

The claim is equivalent to showing that $\tilde{M}_{\pi(i) \pi(j)} = M_{ij}$ for all $i,j$. This can be seen from the construction of SetMask and the definition of $\pi$. But to be more precise, we split the claim into multiple cases: 

Recall that $s(t_i)$ denotes the token sequence containing the token $t_i$ and $q(t_i)$ denotes the set containing $s(t_i)$.

\paragraph{Case 1: $q(t_i) \neq q(t_j)$.} $q(t_i) \neq q(t_j) \iff q(t_{\pi(i)}) \neq q(t_{\pi(j)})$ since $\pi$ does not move tokens between sets. Then by definition of SetMask, $\tilde{M}_{\pi(i) \pi(j)} = 1 = M_{ij}$.

\paragraph{Case 2: $q(t_i) = q(t_j), s(t_i) = s(t_j)$.} Since $\pi$ permutes whole sequences within a set together, all tokens within a sequence remain in the same sequence after permutation. Therefore, $\tilde{M}_{\pi(i) \pi(j)} = 1 = M_{ij}$.

\paragraph{Case 3: $q(t_i) = q(t_j), s(t_i) \neq s(t_j)$.} Again, since $\pi$ permutes whole sequences within a set together, tokens in different sequences of the same set remain in different sequences of the same set after permutation. Therefore, $\tilde{M}_{\pi(i) \pi(j)} = 0 = M_{ij}$.
\qed

\begin{subclaim}
\label{claim:A=PAPT}
    $\tilde{A} = PAP^T$
\end{subclaim}
Again, this claim is equivalent to showing that $\tilde{A}_{\pi(i) \pi(j)} = A_{ij}$ for all $i,j$.
Putting the previous claims together, we have:

\paragraph{Case 1: $M_{ij} = 0$.} By \cref{claim:M=PMPT}, if $M_{ij} = 0$, then $\tilde{M}_{\pi(i) \pi(j)} = 0$ and $A_{ij} = \tilde{A}_{\pi(i) \pi(j)} = 0$.

\paragraph{Case 2: $M_{ij} = 1$.} By \cref{claim:M=PMPT}, if $M_{ij} = 1$, then $\tilde{M}_{\pi(i) \pi(j)} = 1$, and we have
\begin{align}
    \tilde{A}_{\pi(i) \pi(j)}\quad
    &= \frac{exp(\tilde{Z}_{\pi(i) \pi(j)})}{\sum_k \tilde{M}_{\pi(i) \pi(k)} exp(\tilde{Z}_{\pi(i) \pi(k)})} \\
    &\mathrel{\stackrel{\makebox[0pt]{\mbox{\normalfont\tiny \text{(by \cref{claim:M=PMPT,claim:Z=PZPT})}}}}{=}} \qquad \quad \frac{exp(Z_{ij})}{\sum_k M_{ik}exp(Z_{ik})} 
    \quad= A_{ij}
\end{align}
\qed

Finally, putting all the claims together, we have:
\begin{align}
    \tilde{X}^{(t+1)} 
    &= \tilde{A}^{(t)} \tilde{X}^{(t)} W_V \\
    &\mathrel{\stackrel{\makebox[0pt]{\mbox{\normalfont\tiny \text{(by \cref{claim:X=PX,claim:A=PAPT})}}}}{=}} \qquad \quad PA^{(t)}P^T PX^{(t)} W_V \\
    &= PA^{(t)} X^{(t)} W_V \\
    &= P X^{(t+1)},
\end{align}
completing the proof that the attention layer is equivariant.
\end{proofE}

\begin{theoremE}[][end, restate, text link=]
\label{thm:reduction_to_pe}
    Attention with SetPE and SetMask reduces to attention with PE and prefix masking when the input is a single sequence of tokens (i.e., when the input does not contain sets).
\end{theoremE}

\begin{proofE}
If the input is a single sequence of tokens, we have $q = [ q_0 ]$, $q_0 = [ s_0 ]$, and $s_0 = [ \tau_0, \tau_0, \ldots, \tau_{n_{0,0}} ]$. In this case, \cref{alg:setPE} assigns consecutive positions for each token in the input, that is, $[0,1,2, \ldots, n_{0,0}]$. This is identical to absolute positions (PE). 

Similarly, if there are no sets in the input, then SetMask reduces to a fully connected attention mask on the input, i.e., to the prefix mask.
\end{proofE}

Both proofs can be found in \cref{app:proofs}.

\section{Experimental setup}
\label{sec:experimental-setup}

To evaluate the effectiveness of our approach, we test it with $5$ models on $4$ multiple-choice datasets. 
We believe multiple-choice questions are the ideal testbed for Set-LLM since there are widely-available, well-established benchmarks
and they are representative of the scenario where an LLM evaluator selects the best response out of a set of multiple plausible, LLM-generated responses. 






\paragraph{Datasets.}Models are evaluated 
on four popular multiple-choice reasoning benchmarks: PIQA \citep{bisk2019piqa-PIQA}, ARC-Challenge \citep{allenai:arc-ARC}, CommonsenseQA \citep{talmor-etal-2019-commonsenseqa-CSQA}, and SIQA \citep{sap-etal-2019-social-SIQA}. Each task consists of a series of questions, each with multiple choices, where only one answer is correct. 
The number of choices per question varies by the dataset.
In the original setup, PIQA, ARC-Challenge, and CommonsenseQA provide only the question in the prompt, and the different answer choices are run through the model to test them as continuations. 
The choice with the highest log-likelihood is selected as the answer. 

To adapt these datasets to our research setting, we modify the prompt so that the choices are provided as part of the question, and the order of these choices can be permuted. Finally, in the case of SIQA, although the original prompt includes the choices, we remove the numbering since this implicitly assigns an order to the options and breaks the permutation invariance.
All original and modified prompts can be found in \cref{app:prompts}. We use the standard train-evaluation splits for all benchmarks.

\paragraph{Additional pretraining data.}
The Set-LLM adaptations fundamentally change the input of the model and the inner attention mechanism, so the adapted models require training to function in their new setups. To help the models, we experiment with additional pretraining.
We use a high-quality subset (approximately 10k examples) of the cleaned UltraFeedback instruction-following dataset~\citep{cui2024ultrafeedbackboostinglanguagemodels}, attained by following the data preprocessing steps in \citep{kopiczko2024bitune}. 
We use $^{\text{Ultra}}$ in the results to indicate additional pretraining. 
More information on instruction-finetuning can be found in \cref{app:instruction-finetuning}.


\paragraph{Evaluation modes.}
The experiments involve two evaluation modes: \textbf{(1)~Random Order:} 
For each input question, we test all\footnote{\label{footnote:CSQA-24}For CommonsenseQA, only the first $24$ (of a possible $5!=120$) permutations are tested.} permutations of the answer choices, and calculate the average accuracy,
\textbf{(2)~Adversarial Order:} For each input question, we test all\cref{footnote:CSQA-24}
permutations of the answer choices and use a permutation where the LLM returns a wrong answer, if one exists. 
Accuracy is then the proportion of questions for which the LLM remains correct across all possible permutations. 

\paragraph{Base models.}We evaluate the proposed method using 
several popular pretrained decoder-only language models: Gemma 2B and 7B \citep{gemmateam2024gemmaopenmodelsbased-long}, Llama 3.2 1B, Llama 3.2 3B, and Llama 3.1 8B \citep{grattafiori2024llama-short}.\footnote{google/gemma-2b, google/gemma-7b, meta-llama/Llama-3.2-1B-Instruct, meta-llama/Llama-3.2-3B-Instruct, meta-llama/Llama-3.1-8B-Instruct \citep[Huggingface]{wolf2019huggingface}}
We select these models to test different architectures and model sizes.

\paragraph{Baselines.}
We consider the following baselines: 
(1\&2) A pretrained and a finetuned Gemma 2B using the original dataset prompts (Causal Mask+PE$^{*}$ Pretrained \& Causal Mask+PE$^{*}$ Finetuned), 
(3\&4) A pretrained and a finetuned Gemma 2B using the modified dataset prompts (Causal Mask+PE Pretrained \& Causal Mask+PE Finetuned), 
(5) A finetuned Gemma 2B with additional pretraining on UltraFeedback \citep{cui2024ultrafeedbackboostinglanguagemodels} using the modified dataset prompts (Causal Mask+PE$^{\text{Ultra}}$ Finetuned).

An alternative to a permutation-invariant architecture is to run all possible permutations of the input through the LLM and pick the option with the most ``votes''.\footnote{In the event of a tie, the winner is chosen uniformly at random from the top-voted options.} This approach, \textbf{majority vote} \citep{zongfool}, can make any model permutation-invariant.
However, note that this comes with an exponential factor runtime overhead, since the model has to be run $k!$ times, where $k$ is the number of options. 
We include a majority vote for baselines 3, 4, and 5 from above. These models use the modified prompt containing the answer choices, which can be permuted to calculate the majority vote.

\paragraph{Training setup.}
We update the model weights using 
LoRA \citep{hu2022lora}
applied to all linear layers of the multilayer perceptron (MLP) and self-attention layers. 
Details about the hyperparameter settings can be found in \cref{app:hyperparameters}.
We finetune models separately on each benchmark.
We train all models on a single Nvidia H200 GPU with training times ranging between 1 and 4 hours for one model on one benchmark. A comparison of baseline and Set-LLM runtimes is provided in \cref{app:runtimes}.

We train all our models with bfloat16 precision. However, we use full 32-bit floating point precision for all evaluation runs. This proves crucial in ensuring permutation invariance in practice. Although the Set-LLM architecture is provably permutation-invariant (\cref{thm:permutation-invariance}), permuting the input tokens can lead to a different order of the low-level computations resulting in minor inconsistencies, which add up layer by layer. We do not observe any inconsistencies using 32-bit floating-point precision.

\section{Experiments and results}
\label{sec:experiments_and_results}

\subsection{Baselines and order sensitivity}

We first run baseline models and measure the gap between random-order and adversarial-order accuracies. We use Gemma 2B as the base LLM in the first experiments. 



\cref{tab:gemma2B_baselines} shows the results for all Gemma 2B baseline models. 
Finetuning Gemma 2B on the datasets (Causal Mask+PE Finetuned) gives competitive results in the random evaluation mode. However, adversarial permutations lead to large accuracy drops, particularly for ARC-Challenge (55.20\% to 23.72\%).
In contrast, the majority vote baselines do not have any drops in accuracy between the two evaluation modes, confirming that they are permutation-invariant. Moreover, they also produce the best random-order results. However, this comes at a high cost, as the models have to be run $k!$ times for each input. We use Causal Mask+PE$^{\text{Ultra}}$ Finetuned as the baseline for further experiments.

\begin{table}
    \setlength{\tabcolsep}{3.8pt}
    \small
    \caption{
    Gemma 2B baselines on four multiple-choice datasets. 
    All scores are accuracies $(\%)$.
    $^{*}$Results using the original dataset prompts, which for PIQA, ARC, and CSQA only contain the question. All other results use modified prompts with choices provided as part of the question. 
    }
    \label{tab:gemma2B_baselines}
    \centering
    \begin{tabular}{lllc @{\hspace{1\tabcolsep}} cc @{\hspace{1\tabcolsep}} cc @{\hspace{1\tabcolsep}} cc @{\hspace{1\tabcolsep}} c}
         \toprule
         Model & Training & Eval. Mode & \multicolumn{2}{c}{PIQA} & \multicolumn{2}{c}{ARC} & \multicolumn{2}{c}{CSQA} & \multicolumn{2}{c}{SIQA}\\
         & & & Rand. & Adv. & Rand. & Adv. & Rand. & Adv.$^{\dagger}$ & Rand. & Adv. \\
         \midrule
         Random & & - 
         & 50.00 & \textcolor{blue}{50.00}   & 25.00 & \textcolor{blue}{25.00}
         & 20.00 & \textcolor{blue}{20.00}   & 33.33 & \textcolor{blue}{33.33} \\
         \midrule
         Causal Mask+PE$^{*}$ & Pretrained & Single run
         & 76.77 &    & 37.80 &
         & 51.76 &    & 37.26 &  \\
         Causal Mask+PE$^{*}$ & Finetuned & Single run
         & 79.82 &    & 45.39  &
         & 68.80 &    & 75.95 & \\
         \midrule
         Causal Mask+PE & Pretrained & Single run
         & 57.45 & \textcolor{blue}{30.96}   & 36.03 & \textcolor{blue}{\phantom{1}7.68}   
         & 34.92 & \textcolor{blue}{16.46}   & 39.29 & \textcolor{blue}{12.74} \\
         Causal Mask+PE & Pretrained & Majority Vote 
         & 56.04 & \textcolor{blue}{56.04}   & 40.10 & \textcolor{blue}{40.10}   
         & 35.22 & \textcolor{blue}{35.22}   & 40.23 & \textcolor{blue}{40.23} \\

         
         Causal Mask+PE & Finetuned & Single run
         & 84.11 & \textcolor{blue}{76.77}   & 55.20 & \textcolor{blue}{23.72}
         & 78.31 & \textcolor{blue}{69.62}   & 74.80 & \textcolor{blue}{63.00} \\
         Causal Mask+PE & Finetuned & Majority Vote 
         & 84.06 & \textcolor{blue}{84.06}   & 58.87 & \textcolor{blue}{58.87}
         & 78.38 & \textcolor{blue}{78.38}   & 76.05 & \textcolor{blue}{76.05} \\
         Causal Mask+PE$^{\text{Ultra}}$ & Finetuned & Single run
         & 83.98 & \textcolor{blue}{77.31}   & 56.32 & \textcolor{blue}{26.88}
         & 77.89 & \textcolor{blue}{68.47}   & 74.33 & \textcolor{blue}{63.97}\\
         Causal Mask+PE$^{\text{Ultra}}$ & Finetuned & Majority Vote
         & 83.57 & \textcolor{blue}{83.57}   & 59.56 & \textcolor{blue}{59.56}
         & 78.46 & \textcolor{blue}{78.46}   & 75.23 & \textcolor{blue}{75.23} \\
         
         \bottomrule
         \vspace{-0.25cm} \\
         \multicolumn{11}{l}{$^{*}$Results with original prompts 
         \quad $^{\text{Ultra}}$Additional pretraining
         \quad $^{\dagger}$Only first 24 (of 120) permutations tested
         }
    \end{tabular}
\end{table}

\subsection{Set-LLM step-by-step}

As described in \cref{sec:methods}, there are four steps in turning a base LLM into a Set-LLM. To gain a better understanding of the individual steps, we run experiments with the intermediate models: \emph{Causal Mask+NoPE}, \emph{Prefix Mask+NoPE}, \emph{Prefix Mask+SetPE}, and \emph{SetMask+SetPE} (Set-LLM).
Finally, we include \emph{Prefix Mask+PE} in our experiments, which is an encoder-decoder version of the base LLM.
Note that in addition to Set-LLM, Prefix Mask+NoPE and Prefix Mask+SetPE are also set-permutation-invariant models and therefore perform exactly the same in the
two modes.


\cref{tab:gemma_steps} shows the results for all intermediate models.
The set-permutation-invariant models do not have any drops in accuracy in the adversarial setting, confirming our design choices and theoretical results. Moreover, Set-LLM outperforms the strongest baseline on all four benchmarks in both modes, indicating that the permutation-invariance guarantees do not come at a cost to accuracy. This is all the more impressive considering these results come from a single run, rather than $k!$ runs.
Prefix Mask+SetPE is not far off SetMask+SetPE,
but SetMask is needed to outperform the majority vote baseline and give the best results. The problem illustrated in \cref{fig:setPE_fail} might help to explain this gap.
Additional analyses of the majority vote versus Set-LLM output can be found in \cref{app:majority_vote_analysis}.

\begin{table}
    \centering
    \setlength{\tabcolsep}{3.0pt}
    \small
    \caption{
    Set-LLM and all intermediate models going from the base model (Causal Mask+PE) to Set-LLM (SetMask+SetPE). The 4 adaptation steps are described in \cref{sec:methods}. The base model is Gemma 2B. 
    All models are finetuned separately for each benchmark.
    All scores are accuracies $(\%)$.
    }
    \begin{tabular}{lllcc @{\hspace{1\tabcolsep}} cc @{\hspace{1\tabcolsep}} cc @{\hspace{1\tabcolsep}} cc @{\hspace{1\tabcolsep}} c}
         \toprule
         Steps & Model & Eval. Mode & \# Runs & \multicolumn{2}{c}{PIQA} & \multicolumn{2}{c}{ARC} & \multicolumn{2}{c}{CSQA} & \multicolumn{2}{c}{SIQA}\\
         & & & & Rand. & Adv. & Rand. & Adv. & Rand. & Adv.$^{\dagger}$ & Rand. & Adv. \\
         \midrule
         - & Causal Mask+PE$^{\text{Ultra}}$ & Majority Vote & $k!$
         & 83.57 & \textcolor{blue}{83.57}   & 59.56 & \textcolor{blue}{59.56}
         & 78.46 & \textcolor{blue}{78.46}   & 75.23 & \textcolor{blue}{75.23} \\
         \midrule
         
         - & Causal Mask+PE & Single run & 1
         & 84.11 & \textcolor{blue}{76.77}   & 55.20 & \textcolor{blue}{23.72}
         & 78.31 & \textcolor{blue}{69.62}   & 74.80 & \textcolor{blue}{63.00} \\
         
         1 & Causal Mask+NoPE & Single run & 1
         & 74.37 & \textcolor{blue}{63.55}   & 35.70 & \textcolor{blue}{14.76}
         & 68.49 & \textcolor{blue}{57.33}   & 63.21 & \textcolor{blue}{48.57} \\
         1,2 & Prefix Mask+NoPE & Single run & 1
         & 74.54 & \textcolor{blue}{74.54}   & 32.08 & \textcolor{blue}{32.08}
         & 49.14 & \textcolor{blue}{49.14}   & 51.02 & \textcolor{blue}{51.02} \\
         2 & Prefix Mask+PE & Single run & 1
         & 82.78 & \textcolor{blue}{76.50}   & 57.62 & \textcolor{blue}{27.47}
         & 78.98 & \textcolor{blue}{71.01}   & 74.36 & \textcolor{blue}{65.66} \\
         1-3 & Prefix Mask+SetPE & Single run & 1
         & 81.23 & \textcolor{blue}{81.23}   & 51.28 & \textcolor{blue}{51.28}
         & 77.31 & \textcolor{blue}{77.31}   & 71.24 & \textcolor{blue}{71.24} \\
         1-4 & SetMask+SetPE & Single run & 1
         & 84.33 & \textcolor{blue}{84.33}   & 57.76 & \textcolor{blue}{57.76}
         & 79.93 & \textcolor{blue}{79.93}   & 75.38 & \textcolor{blue}{75.38} \\         
         1-4 & SetMask+SetPE$^{\text{Ultra}}$ & Single run & 1
         & \textbf{85.80} & \textbf{\textcolor{blue}{85.80}}   & \textbf{65.02} & \textbf{\textcolor{blue}{65.02}}
         & \textbf{80.18} & \textbf{\textcolor{blue}{80.18}}   & \textbf{76.15} & \textbf{\textcolor{blue}{76.15}} \\
         \bottomrule
         \vspace{-0.25cm} \\
         \multicolumn{12}{l}{
         $^{\text{Ultra}}$Additional pretraining
         \quad $^{\dagger}$Only first 24 (of 120) permutations tested
         \quad $k=\text{number of (multiple) choices}$
         } 
    \end{tabular}
    \label{tab:gemma_steps}
\end{table}

\subsection{Different base LLMs}

In addition to Gemma 2B, we evaluate Set-LLM using Gemma 7B, Llama 3.2 1B, Llama 3.2 3B, and Llama 3.1 8B as base models. \cref{tab:all_models} shows that all base models suffer from order sensitivity with drops between the two evaluation modes ranging from $3.4\%$ to $31.7\%$. In contrast, there are no drops with Set-LLM, and Set-LLM outperforms the base model in 20/20 cases with adversarial ordering and in 18/20 cases with random ordering. Moreover, with a single run, it outperforms majority vote in 16/20 cases, without the exponential runtime overhead. 
Set-LLM outperforms the baselines across all model architectures and sizes we tested.

\begin{table}
    \centering
    \setlength{\tabcolsep}{3.1pt}
    \small
    \caption{
    Set-LLM performance with different base LLMs. All models were pretrained on UltraFeedback \citep{cui2024ultrafeedbackboostinglanguagemodels}
    and then finetuned separately for each benchmark.
    All scores are accuracies $(\%)$.
    }
    \begin{tabular}{lllc @{\hspace{1\tabcolsep}} cc @{\hspace{1\tabcolsep}} cc @{\hspace{1\tabcolsep}} cc @{\hspace{1\tabcolsep}} c}
         \toprule
         LLM & Model & Eval. Mode & \multicolumn{2}{c}{PIQA} & \multicolumn{2}{c}{ARC} & \multicolumn{2}{c}{CSQA} & \multicolumn{2}{c}{SIQA}\\
         & & & Rand. & Adv. & Rand. & Adv. & Rand. & Adv.$^{\dagger}$ & Rand. & Adv. \\
         \midrule
         \multirow{3}{*}{Gemma 2B}
         & Causal Mask+PE$^{\text{Ultra}}$ & Single run
         & 83.98 & \textcolor{blue}{77.31}   & 56.32 & \textcolor{blue}{26.88}
         & 77.89 & \textcolor{blue}{68.47}   & 74.33 & \textcolor{blue}{63.97}\\
         & Causal Mask+PE$^{\text{Ultra}}$ & Majority Vote
         & 84.17 & \textcolor{blue}{84.17}   & 60.15 & \textcolor{blue}{60.15}
         & 78.71 & \textcolor{blue}{78.71}   & 75.38 & \textcolor{blue}{75.38} \\
         & SetMask+SetPE$^{\text{Ultra}}$ & Single run
         & \textbf{85.80} & \textbf{\textcolor{blue}{85.80}}   & \textbf{65.02} & \textbf{\textcolor{blue}{65.02}}
         & \textbf{80.18} & \textbf{\textcolor{blue}{80.18}}   & \textbf{76.15} & \textbf{\textcolor{blue}{76.15}} \\
         \midrule
         \multirow{3}{*}{Gemma 7B}
         & Causal Mask+PE$^{\text{Ultra}}$ & Single run
         & 92.82 & \textcolor{blue}{89.45}   & 83.52 & \textcolor{blue}{64.33}
         & 85.45 & \textcolor{blue}{79.12}   & 80.93 & \textcolor{blue}{74.10} \\
         & Causal Mask+PE$^{\text{Ultra}}$ & Majority Vote
         & 92.66 & \textcolor{blue}{92.66}   & \textbf{85.58} & \textbf{\textcolor{blue}{85.58}}
         & \textbf{85.75} & \textbf{\textcolor{blue}{85.75}}   & 81.10 & \textcolor{blue}{81.10} \\
         & SetMask+SetPE$^{\text{Ultra}}$ & Single run
         & \textbf{92.98} & \textbf{\textcolor{blue}{92.98}}   & 83.45 & \textcolor{blue}{83.45}
         & 84.93 & \textcolor{blue}{84.93}   & \textbf{81.12} & \textbf{\textcolor{blue}{81.12}} \\
         \midrule
         \multirow{3}{*}{Llama 3.2 1B}
         & Causal Mask+PE$^{\text{Ultra}}$ & Single run
         & 79.57 & \textcolor{blue}{71.33}   & 53.61 & \textcolor{blue}{21.93}
         & 74.50 & \textcolor{blue}{64.21}   & 71.84 & \textcolor{blue}{62.79} \\
         & Causal Mask+PE$^{\text{Ultra}}$ & Majority Vote
         & 79.49 & \textcolor{blue}{79.49}   & 57.17 & \textcolor{blue}{57.17}
         & 75.51 & \textcolor{blue}{75.51}   & 71.85 & \textcolor{blue}{71.85} \\
         & SetMask+SetPE$^{\text{Ultra}}$ & Single run
         & \textbf{81.66} & \textbf{\textcolor{blue}{81.66}}   & \textbf{59.30} & \textbf{\textcolor{blue}{59.30}}
         & \textbf{76.66} & \textbf{\textcolor{blue}{76.66}}   & \textbf{72.47} & \textbf{\textcolor{blue}{72.47}} \\
         \midrule
         \multirow{3}{*}{Llama 3.2 3B}
         & Causal Mask+PE$^{\text{Ultra}}$ & Single run
         & 86.92 & \textcolor{blue}{81.72}   & 74.16 & \textcolor{blue}{53.07}
         & 81.32 & \textcolor{blue}{74.94}   & 77.54 & \textcolor{blue}{70.42} \\
         & Causal Mask+PE$^{\text{Ultra}}$ & Majority Vote
         & 86.83 & \textcolor{blue}{86.83}   & \textbf{76.37} & \textbf{\textcolor{blue}{76.37}}
         & 81.57 & \textcolor{blue}{81.57}   & 77.99 & \textcolor{blue}{77.99} \\
         & SetMask+SetPE$^{\text{Ultra}}$ & Single run
         & \textbf{88.41} & \textbf{\textcolor{blue}{88.41}}   & 75.85 & \textcolor{blue}{75.85}
         & \textbf{83.29} & \textbf{\textcolor{blue}{83.29}}   & \textbf{80.30} & \textbf{\textcolor{blue}{80.30}} \\
         \midrule
         \multirow{3}{*}{Llama 3.1 8B}
         & Causal Mask+PE$^{\text{Ultra}}$ & Single run
         & 90.81 & \textcolor{blue}{86.29}   & 83.04 & \textcolor{blue}{64.51}   
         & 83.96 & \textcolor{blue}{77.89}   & 80.77 & \textcolor{blue}{73.90} \\
         & Causal Mask+PE$^{\text{Ultra}}$ & Majority Vote
         & 90.75 & \textcolor{blue}{90.75}   & \textbf{85.32} & \textbf{\textcolor{blue}{85.32}} 
         & 84.11 & \textcolor{blue}{84.11}   & 81.12 & \textcolor{blue}{81.12} \\
         & SetMask+SetPE$^{\text{Ultra}}$ & Single run
         & \textbf{91.62} & \textbf{\textcolor{blue}{91.62}}   & 84.13 & \textcolor{blue}{84.13} 
         & \textbf{85.34} & \textbf{\textcolor{blue}{85.34}}   & \textbf{81.47} & \textbf{\textcolor{blue}{81.47}} \\
         \bottomrule
         \vspace{-0.25cm} \\
         \multicolumn{8}{l}{
         $^{\text{Ultra}}$Additional pretraining
         \quad $^{\dagger}$Only first 24 (of 120) permutations tested
         } 
    \end{tabular}
    \label{tab:all_models}
\end{table}

\subsection{Out-of-distribution performance}

When using Set-LLM as an LLM evaluator, it is particularly important that the adaptations do not hurt the performance on out-of-distribution data after finetuning on a small dataset. 
We measure the out-of-distribution multiple-choice performance of the models by finetuning a model on each benchmark independently and then evaluating on the three remaining benchmarks. We do this for both the base model and Set-LLM and compare their performance with the pretrained base model.
Results are provided in \cref{tab:ood} in the Appendix. Set-LLM is the best-performing model in 10/12 cases in both evaluation modes.

\section{Related work}

To the best of our knowledge, this is the first paper to specifically introduce a permutation-invariant architecture for decoder-only language models. However, there are many relevant works that this paper builds on as well as works that combine text and permutation-invariant inputs in one model.

\paragraph{Permuting multiple-choice questions.} 
This work was motivated by the observations that you can ``fool your (vision and) language model with embarrassingly simple permutations'' \citep{zongfool}. The authors quantify the effect of adversarial permutations on (V)LLMs and multiple-choice benchmarks. They also analyze the effectiveness of majority vote, which we include as a baseline in our experiments. 
Prior to this, \citet{liu2024mmbench-circular-eval} suggested rotating the choices of multiple-choice questions to evaluate the robustness of (Multimodal) Language Models. 
Taking a majority vote over the rotations could be used as an alternative to majority vote, thereby only carrying a linear-factor rather than an exponential-factor runtime overhead, but it is not guaranteed to give the same answer for all permutations.

\paragraph{LLM evaluators.} LLM evaluators (or judges) were introduced recently \citep{zheng2023judgingLLM-as-a-Judge-MT-Arena, zhu2023judgelm}, but have already been integrated in many LLM pipelines, including for evaluation, retrieval and reasoning \citep{gu2024survey-on-llm-as-a-judge, badshah2024reference-LLMs-as-Judges-for-Free-Form-text, gao2024strategyllm-LLM-judge-for-reasoning, liang-etal-2024-encouraging-LLM-judge-for-reasoning}. The order bias of LLM evaluators was already observed by \citet{zheng2023judgingLLM-as-a-Judge-MT-Arena}, and methods involving the aggregation of multiple runs and specialized prompting have been proposed to mitigate this problem \citep{liu2024mmbench-circular-eval, zongfool, kim-etal-2025-biggen-llm-judge, zhuang2024setwise-prompting, tang2023found-fixed-ordered-permutations-for-self-consistency}. 
However, this is the first work to eliminate order bias directly from the model architecture without impacting accuracy or runtime. 


\paragraph{Graphs and large language models.} 
A set can be seen as an empty (or fully-connected) graph, and indeed one of the key features of graphs is also permutation-invariance. As a result, several works that combine graphs and LLMs are also relevant here \citep{yoon2023multimodal-graph-learning, he2024G-retriever, yang2021graphformers}.
Most closely related to this work are the papers of \citet{liu2020k-bert-graph-llm} and \citet{plenz-frank-2024-graphLMs}, who adapt an encoder-decoder and an encoder-only language model, respectively, to take mixed graph-text data as input. Both approaches alter the attention mechanism and the positional encoding to incorporate graph connectivity information into the embeddings.

\paragraph{Invariant neural networks.}While invariant networks have existed for some time, \citet{bronstein2021geometric} recently unified them under a common geometric framework. For example, CNNs \citep{zhang1988shift-invariant-CNN, lecun1989backpropagation-CNN} are shift-invariant architectures for images, Deep Sets \citep{zaheer2017deepsets} was the first permutation-invariant architecture for sets, and GNNs \citep{kipf2016semi-GCN, velivckovic2017graph-GAT, xu2018powerful-GIN} are permutation-invariant architectures for graphs. In this work, we combine the permutation invariance of these architectures with the power of large language models.

\section{Conclusion}
\label{sec:conclusion}

To the best of our knowledge, this paper introduces the first permutation-invariant decoder-only LLM, Set-LLM. We formally prove that Set-LLM is permutation-invariant and show how robust it is to permutations in practice. 
The models are based on pretrained LLMs, can be finetuned efficiently with low-rank adapters, and incur no additional runtime costs. 
We test Set-LLM on four multiple-choice datasets, where it outperforms all baselines by significant margins in either accuracy or runtime.

With the growing importance of LLMs as evaluators, Set-LLM has the potential for wider impact. It can improve the reliability of automated evaluation metrics and improve the decision-making of reasoning LLMs 
for solving complex problems. 
Our approach could also apply to other mixed set-text scenarios, such as generation with a set of supporting documents (as in retrieval-augmented generation) or graph reasoning, with the graph represented as a set of edges. 

\paragraph{Limitations.}
\label{para:limitations}
While we propose a general purpose set-permutation-invariant LLM, the experiments focus on multiple-choice question answering. We believe this is the ideal test application for the proposed approach since there are widely-available, well-established benchmarks and standardized evaluation frameworks in this area. 



\makeatletter
\if@preprint
\section*{Acknowledgments}

The authors would like to thank
the Vector Stiftung for their financial support in the framework of the MINT Innovations program.
\fi



\bibliography{references}

\begin{thebibliography}{50}
\providecommand{\natexlab}[1]{#1}
\providecommand{\url}[1]{\texttt{#1}}
\expandafter\ifx\csname urlstyle\endcsname\relax
  \providecommand{\doi}[1]{doi: #1}\else
  \providecommand{\doi}{doi: \begingroup \urlstyle{rm}\Url}\fi

\bibitem[Achiam et~al.(2023)Achiam, Adler, Agarwal, Ahmad, Akkaya, Aleman, Almeida, Altenschmidt, Altman, Anadkat, et~al.]{achiam2023gpt}
Josh Achiam, Steven Adler, Sandhini Agarwal, Lama Ahmad, Ilge Akkaya, Florencia~Leoni Aleman, Diogo Almeida, Janko Altenschmidt, Sam Altman, Shyamal Anadkat, et~al.
\newblock Gpt-4 technical report.
\newblock \emph{arXiv preprint arXiv:2303.08774}, 2023.

\bibitem[Badshah and Sajjad(2024)]{badshah2024reference-LLMs-as-Judges-for-Free-Form-text}
Sher Badshah and Hassan Sajjad.
\newblock Reference-guided verdict: Llms-as-judges in automatic evaluation of free-form text.
\newblock \emph{arXiv preprint arXiv:2408.09235}, 2024.

\bibitem[Bisk et~al.(2019)Bisk, Zellers, Gao, and Choi]{bisk2019piqa-PIQA}
Yonatan Bisk, Rowan Zellers, J~Gao, and Y~Choi.
\newblock Piqa: reasoning about physical commonsense in natural language. corr, vol. abs/1911.11641, 2019.

\bibitem[Bronstein et~al.(2021)Bronstein, Bruna, Cohen, and Veli{\v{c}}kovi{\'c}]{bronstein2021geometric}
Michael~M Bronstein, Joan Bruna, Taco Cohen, and Petar Veli{\v{c}}kovi{\'c}.
\newblock Geometric deep learning: Grids, groups, graphs, geodesics, and gauges.
\newblock \emph{arXiv preprint arXiv:2104.13478}, 2021.

\bibitem[Clark et~al.(2018)Clark, Cowhey, Etzioni, Khot, Sabharwal, Schoenick, and Tafjord]{allenai:arc-ARC}
Peter Clark, Isaac Cowhey, Oren Etzioni, Tushar Khot, Ashish Sabharwal, Carissa Schoenick, and Oyvind Tafjord.
\newblock Think you have solved question answering? try arc, the ai2 reasoning challenge.
\newblock \emph{arXiv:1803.05457v1}, 2018.

\bibitem[Cui et~al.(2024)Cui, Yuan, Ding, Yao, He, Zhu, Ni, Xie, Xie, Lin, Liu, and Sun]{cui2024ultrafeedbackboostinglanguagemodels}
Ganqu Cui, Lifan Yuan, Ning Ding, Guanming Yao, Bingxiang He, Wei Zhu, Yuan Ni, Guotong Xie, Ruobing Xie, Yankai Lin, Zhiyuan Liu, and Maosong Sun.
\newblock Ultrafeedback: Boosting language models with scaled ai feedback, 2024.
\newblock URL \url{https://arxiv.org/abs/2310.01377}.

\bibitem[Dubois et~al.(2023)Dubois, Li, Taori, Zhang, Gulrajani, Ba, Guestrin, Liang, and Hashimoto]{dubois2023alpacafarm-judge-order-bias}
Yann Dubois, Chen~Xuechen Li, Rohan Taori, Tianyi Zhang, Ishaan Gulrajani, Jimmy Ba, Carlos Guestrin, Percy~S Liang, and Tatsunori~B Hashimoto.
\newblock Alpacafarm: A simulation framework for methods that learn from human feedback.
\newblock \emph{Advances in Neural Information Processing Systems}, 36:\penalty0 30039--30069, 2023.

\bibitem[Gao et~al.(2024)Gao, Jiang, Cai, Shi, and Lam]{gao2024strategyllm-LLM-judge-for-reasoning}
Chang Gao, Haiyun Jiang, Deng Cai, Shuming Shi, and Wai Lam.
\newblock Strategyllm: Large language models as strategy generators, executors, optimizers, and evaluators for problem solving.
\newblock \emph{Advances in Neural Information Processing Systems}, 37:\penalty0 96797--96846, 2024.

\bibitem[Grattafiori et~al.(2024)Grattafiori, Dubey, Jauhri, Pandey, Kadian, Al-Dahle, Letman, Mathur, Schelten, Vaughan, et~al.]{grattafiori2024llama-short}
Aaron Grattafiori, Abhimanyu Dubey, Abhinav Jauhri, Abhinav Pandey, Abhishek Kadian, Ahmad Al-Dahle, Aiesha Letman, Akhil Mathur, Alan Schelten, Alex Vaughan, et~al.
\newblock The llama 3 herd of models.
\newblock \emph{arXiv preprint arXiv:2407.21783}, 2024.

\bibitem[Gu et~al.(2024)Gu, Jiang, Shi, Tan, Zhai, Xu, Li, Shen, Ma, Liu, et~al.]{gu2024survey-on-llm-as-a-judge}
Jiawei Gu, Xuhui Jiang, Zhichao Shi, Hexiang Tan, Xuehao Zhai, Chengjin Xu, Wei Li, Yinghan Shen, Shengjie Ma, Honghao Liu, et~al.
\newblock A survey on llm-as-a-judge.
\newblock \emph{arXiv preprint arXiv:2411.15594}, 2024.

\bibitem[Guo et~al.(2021)Guo, Sablayrolles, J{\'e}gou, and Kiela]{guo-etal-2021-gradient-adversarial-attacks}
Chuan Guo, Alexandre Sablayrolles, Herv{\'e} J{\'e}gou, and Douwe Kiela.
\newblock Gradient-based adversarial attacks against text transformers.
\newblock In Marie-Francine Moens, Xuanjing Huang, Lucia Specia, and Scott Wen-tau Yih, editors, \emph{Proceedings of the 2021 Conference on Empirical Methods in Natural Language Processing}, pages 5747--5757, Online and Punta Cana, Dominican Republic, November 2021. Association for Computational Linguistics.
\newblock \doi{10.18653/v1/2021.emnlp-main.464}.
\newblock URL \url{https://aclanthology.org/2021.emnlp-main.464/}.

\bibitem[He et~al.(2024)He, Tian, Sun, Chawla, Laurent, LeCun, Bresson, and Hooi]{he2024G-retriever}
Xiaoxin He, Yijun Tian, Yifei Sun, Nitesh Chawla, Thomas Laurent, Yann LeCun, Xavier Bresson, and Bryan Hooi.
\newblock G-retriever: Retrieval-augmented generation for textual graph understanding and question answering.
\newblock \emph{Advances in Neural Information Processing Systems}, 37:\penalty0 132876--132907, 2024.

\bibitem[Hu et~al.(2022)Hu, Shen, Wallis, Allen-Zhu, Li, Wang, Wang, and Chen]{hu2022lora}
Edward~J Hu, Yelong Shen, Phillip Wallis, Zeyuan Allen-Zhu, Yuanzhi Li, Shean Wang, Lu~Wang, and Weizhu Chen.
\newblock Lo{RA}: Low-rank adaptation of large language models.
\newblock In \emph{International Conference on Learning Representations}, 2022.
\newblock URL \url{https://openreview.net/forum?id=nZeVKeeFYf9}.

\bibitem[Jiang et~al.(2023)Jiang, Sablayrolles, Mensch, Bamford, Chaplot, Casas, Bressand, Lengyel, Lample, Saulnier, et~al.]{jiang2023mistral}
Albert~Q Jiang, A~Sablayrolles, A~Mensch, C~Bamford, D~Singh Chaplot, Ddl Casas, F~Bressand, G~Lengyel, G~Lample, L~Saulnier, et~al.
\newblock Mistral 7b. arxiv.
\newblock \emph{arXiv preprint arXiv:2310.06825}, 10, 2023.

\bibitem[Kazemnejad et~al.(2023)Kazemnejad, Padhi, Natesan~Ramamurthy, Das, and Reddy]{kazemnejad2023impactNoPE}
Amirhossein Kazemnejad, Inkit Padhi, Karthikeyan Natesan~Ramamurthy, Payel Das, and Siva Reddy.
\newblock The impact of positional encoding on length generalization in transformers.
\newblock \emph{Advances in Neural Information Processing Systems}, 36:\penalty0 24892--24928, 2023.

\bibitem[Kim et~al.(2025)Kim, Suk, Cho, Longpre, Kim, Yoon, Son, Cho, Shafayat, Baek, Park, Hwang, Jo, Cho, Shin, Lee, Oh, Lee, Ho, Joo, Ko, Lee, Chae, Shin, Jang, Ye, Lin, Welleck, Neubig, Lee, Lee, and Seo]{kim-etal-2025-biggen-llm-judge}
Seungone Kim, Juyoung Suk, Ji~Yong Cho, Shayne Longpre, Chaeeun Kim, Dongkeun Yoon, Guijin Son, Yejin Cho, Sheikh Shafayat, Jinheon Baek, Sue~Hyun Park, Hyeonbin Hwang, Jinkyung Jo, Hyowon Cho, Haebin Shin, Seongyun Lee, Hanseok Oh, Noah Lee, Namgyu Ho, Se~June Joo, Miyoung Ko, Yoonjoo Lee, Hyungjoo Chae, Jamin Shin, Joel Jang, Seonghyeon Ye, Bill~Yuchen Lin, Sean Welleck, Graham Neubig, Moontae Lee, Kyungjae Lee, and Minjoon Seo.
\newblock The {B}i{GG}en bench: A principled benchmark for fine-grained evaluation of language models with language models.
\newblock In Luis Chiruzzo, Alan Ritter, and Lu~Wang, editors, \emph{Proceedings of the 2025 Conference of the Nations of the Americas Chapter of the Association for Computational Linguistics: Human Language Technologies (Volume 1: Long Papers)}, pages 5877--5919, Albuquerque, New Mexico, April 2025. Association for Computational Linguistics.
\newblock ISBN 979-8-89176-189-6.
\newblock URL \url{https://aclanthology.org/2025.naacl-long.303/}.

\bibitem[Kipf and Welling(2016)]{kipf2016semi-GCN}
Thomas~N Kipf and Max Welling.
\newblock Semi-supervised classification with graph convolutional networks.
\newblock \emph{arXiv preprint arXiv:1609.02907}, 2016.

\bibitem[Kopiczko et~al.(2024)Kopiczko, Blankevoort, and Asano]{kopiczko2024bitune}
Dawid~J Kopiczko, Tijmen Blankevoort, and Yuki~M Asano.
\newblock Bitune: Bidirectional instruction-tuning.
\newblock \emph{arXiv preprint arXiv:2405.14862}, 2024.

\bibitem[LeCun et~al.(1989)LeCun, Boser, Denker, Henderson, Howard, Hubbard, and Jackel]{lecun1989backpropagation-CNN}
Yann LeCun, Bernhard Boser, John~S Denker, Donnie Henderson, Richard~E Howard, Wayne Hubbard, and Lawrence~D Jackel.
\newblock Backpropagation applied to handwritten zip code recognition.
\newblock \emph{Neural computation}, 1\penalty0 (4):\penalty0 541--551, 1989.

\bibitem[Lhoest et~al.(2021)Lhoest, Villanova~del Moral, Jernite, Thakur, von Platen, Patil, Chaumond, Drame, Plu, Tunstall, Davison, {\v{S}}a{\v{s}}ko, Chhablani, Malik, Brandeis, Le~Scao, Sanh, Xu, Patry, McMillan-Major, Schmid, Gugger, Delangue, Matussi{\`e}re, Debut, Bekman, Cistac, Goehringer, Mustar, Lagunas, Rush, and Wolf]{lhoest-etal-2021-datasets-huggingface}
Quentin Lhoest, Albert Villanova~del Moral, Yacine Jernite, Abhishek Thakur, Patrick von Platen, Suraj Patil, Julien Chaumond, Mariama Drame, Julien Plu, Lewis Tunstall, Joe Davison, Mario {\v{S}}a{\v{s}}ko, Gunjan Chhablani, Bhavitvya Malik, Simon Brandeis, Teven Le~Scao, Victor Sanh, Canwen Xu, Nicolas Patry, Angelina McMillan-Major, Philipp Schmid, Sylvain Gugger, Cl{\'e}ment Delangue, Th{\'e}o Matussi{\`e}re, Lysandre Debut, Stas Bekman, Pierric Cistac, Thibault Goehringer, Victor Mustar, Fran{\c{c}}ois Lagunas, Alexander Rush, and Thomas Wolf.
\newblock Datasets: A community library for natural language processing.
\newblock In Heike Adel and Shuming Shi, editors, \emph{Proceedings of the 2021 Conference on Empirical Methods in Natural Language Processing: System Demonstrations}, pages 175--184, Online and Punta Cana, Dominican Republic, November 2021. Association for Computational Linguistics.
\newblock \doi{10.18653/v1/2021.emnlp-demo.21}.
\newblock URL \url{https://aclanthology.org/2021.emnlp-demo.21/}.

\bibitem[Li et~al.(2023)Li, Wang, Ding, and Chen]{li2023llminfinancesurvey}
Yinheng Li, Shaofei Wang, Han Ding, and Hang Chen.
\newblock Large language models in finance: A survey.
\newblock In \emph{Proceedings of the fourth ACM international conference on AI in finance}, pages 374--382, 2023.

\bibitem[Liang et~al.(2024)Liang, He, Jiao, Wang, Wang, Wang, Yang, Shi, and Tu]{liang-etal-2024-encouraging-LLM-judge-for-reasoning}
Tian Liang, Zhiwei He, Wenxiang Jiao, Xing Wang, Yan Wang, Rui Wang, Yujiu Yang, Shuming Shi, and Zhaopeng Tu.
\newblock Encouraging divergent thinking in large language models through multi-agent debate.
\newblock In Yaser Al-Onaizan, Mohit Bansal, and Yun-Nung Chen, editors, \emph{Proceedings of the 2024 Conference on Empirical Methods in Natural Language Processing}, pages 17889--17904, Miami, Florida, USA, November 2024. Association for Computational Linguistics.
\newblock \doi{10.18653/v1/2024.emnlp-main.992}.
\newblock URL \url{https://aclanthology.org/2024.emnlp-main.992/}.

\bibitem[Liu et~al.(2020)Liu, Zhou, Zhao, Wang, Ju, Deng, and Wang]{liu2020k-bert-graph-llm}
Weijie Liu, Peng Zhou, Zhe Zhao, Zhiruo Wang, Qi~Ju, Haotang Deng, and Ping Wang.
\newblock K-bert: Enabling language representation with knowledge graph.
\newblock In \emph{Proceedings of the AAAI conference on artificial intelligence}, volume~34, pages 2901--2908, 2020.

\bibitem[Liu et~al.(2024)Liu, Duan, Zhang, Li, Zhang, Zhao, Yuan, Wang, He, Liu, et~al.]{liu2024mmbench-circular-eval}
Yuan Liu, Haodong Duan, Yuanhan Zhang, Bo~Li, Songyang Zhang, Wangbo Zhao, Yike Yuan, Jiaqi Wang, Conghui He, Ziwei Liu, et~al.
\newblock Mmbench: Is your multi-modal model an all-around player?
\newblock In \emph{European conference on computer vision}, pages 216--233. Springer, 2024.

\bibitem[Mangrulkar et~al.(2022)Mangrulkar, Gugger, Debut, Belkada, Paul, and Bossan]{2022peft-huggingface}
Sourab Mangrulkar, Sylvain Gugger, Lysandre Debut, Younes Belkada, Sayak Paul, and Benjamin Bossan.
\newblock Peft: State-of-the-art parameter-efficient fine-tuning methods.
\newblock \url{https://github.com/huggingface/peft}, 2022.

\bibitem[Ouyang et~al.(2022)Ouyang, Wu, Jiang, Almeida, Wainwright, Mishkin, Zhang, Agarwal, Slama, Ray, et~al.]{ouyang2022instructiontuning}
Long Ouyang, Jeffrey Wu, Xu~Jiang, Diogo Almeida, Carroll Wainwright, Pamela Mishkin, Chong Zhang, Sandhini Agarwal, Katarina Slama, Alex Ray, et~al.
\newblock Training language models to follow instructions with human feedback.
\newblock \emph{Advances in neural information processing systems}, 35:\penalty0 27730--27744, 2022.

\bibitem[Plenz and Frank(2024)]{plenz-frank-2024-graphLMs}
Moritz Plenz and Anette Frank.
\newblock Graph language models.
\newblock In Lun-Wei Ku, Andre Martins, and Vivek Srikumar, editors, \emph{Proceedings of the 62nd Annual Meeting of the Association for Computational Linguistics (Volume 1: Long Papers)}, pages 4477--4494, Bangkok, Thailand, August 2024. Association for Computational Linguistics.
\newblock \doi{10.18653/v1/2024.acl-long.245}.
\newblock URL \url{https://aclanthology.org/2024.acl-long.245/}.

\bibitem[Raffel et~al.(2020)Raffel, Shazeer, Roberts, Lee, Narang, Matena, Zhou, Li, and Liu]{raffel2020exploringT5}
Colin Raffel, Noam Shazeer, Adam Roberts, Katherine Lee, Sharan Narang, Michael Matena, Yanqi Zhou, Wei Li, and Peter~J Liu.
\newblock Exploring the limits of transfer learning with a unified text-to-text transformer.
\newblock \emph{Journal of machine learning research}, 21\penalty0 (140):\penalty0 1--67, 2020.

\bibitem[Sap et~al.(2019)Sap, Rashkin, Chen, Le~Bras, and Choi]{sap-etal-2019-social-SIQA}
Maarten Sap, Hannah Rashkin, Derek Chen, Ronan Le~Bras, and Yejin Choi.
\newblock Social {IQ}a: Commonsense reasoning about social interactions.
\newblock In Kentaro Inui, Jing Jiang, Vincent Ng, and Xiaojun Wan, editors, \emph{Proceedings of the 2019 Conference on Empirical Methods in Natural Language Processing and the 9th International Joint Conference on Natural Language Processing (EMNLP-IJCNLP)}, pages 4463--4473, Hong Kong, China, November 2019. Association for Computational Linguistics.
\newblock \doi{10.18653/v1/D19-1454}.
\newblock URL \url{https://aclanthology.org/D19-1454/}.

\bibitem[Shaw et~al.(2018)Shaw, Uszkoreit, and Vaswani]{shaw-etal-2018-self-relative-PE}
Peter Shaw, Jakob Uszkoreit, and Ashish Vaswani.
\newblock Self-attention with relative position representations.
\newblock In Marilyn Walker, Heng Ji, and Amanda Stent, editors, \emph{Proceedings of the 2018 Conference of the North {A}merican Chapter of the Association for Computational Linguistics: Human Language Technologies, Volume 2 (Short Papers)}, pages 464--468, New Orleans, Louisiana, June 2018. Association for Computational Linguistics.
\newblock \doi{10.18653/v1/N18-2074}.
\newblock URL \url{https://aclanthology.org/N18-2074/}.

\bibitem[Shayegani et~al.(2023)Shayegani, Mamun, Fu, Zaree, Dong, and Abu-Ghazaleh]{shayegani2023-llm-vulnerabilities-survey}
Erfan Shayegani, Md~Abdullah~Al Mamun, Yu~Fu, Pedram Zaree, Yue Dong, and Nael Abu-Ghazaleh.
\newblock Survey of vulnerabilities in large language models revealed by adversarial attacks.
\newblock \emph{arXiv preprint arXiv:2310.10844}, 2023.

\bibitem[Su et~al.(2024)Su, Ahmed, Lu, Pan, Bo, and Liu]{su2024roformer-ROPE}
Jianlin Su, Murtadha Ahmed, Yu~Lu, Shengfeng Pan, Wen Bo, and Yunfeng Liu.
\newblock Roformer: Enhanced transformer with rotary position embedding.
\newblock \emph{Neurocomputing}, 568:\penalty0 127063, 2024.

\bibitem[Talmor et~al.(2019)Talmor, Herzig, Lourie, and Berant]{talmor-etal-2019-commonsenseqa-CSQA}
Alon Talmor, Jonathan Herzig, Nicholas Lourie, and Jonathan Berant.
\newblock {C}ommonsense{QA}: A question answering challenge targeting commonsense knowledge.
\newblock In Jill Burstein, Christy Doran, and Thamar Solorio, editors, \emph{Proceedings of the 2019 Conference of the North {A}merican Chapter of the Association for Computational Linguistics: Human Language Technologies, Volume 1 (Long and Short Papers)}, pages 4149--4158, Minneapolis, Minnesota, June 2019. Association for Computational Linguistics.
\newblock \doi{10.18653/v1/N19-1421}.
\newblock URL \url{https://aclanthology.org/N19-1421/}.

\bibitem[Tang et~al.(2023)Tang, Zhang, Ma, Lin, and Ture]{tang2023found-fixed-ordered-permutations-for-self-consistency}
Raphael Tang, Xinyu Zhang, Xueguang Ma, Jimmy Lin, and Ferhan Ture.
\newblock Found in the middle: Permutation self-consistency improves listwise ranking in large language models.
\newblock \emph{arXiv preprint arXiv:2310.07712}, 2023.

\bibitem[Team et~al.(2024)Team, Mesnard, Hardin, Dadashi, Bhupatiraju, Pathak, Sifre, Rivière, Kale, Love, Tafti, Hussenot, Sessa, Chowdhery, Roberts, Barua, Botev, Castro-Ros, Slone, Héliou, Tacchetti, Bulanova, Paterson, Tsai, Shahriari, Lan, Choquette-Choo, Crepy, Cer, Ippolito, Reid, Buchatskaya, Ni, Noland, Yan, Tucker, Muraru, Rozhdestvenskiy, Michalewski, Tenney, Grishchenko, Austin, Keeling, Labanowski, Lespiau, Stanway, Brennan, Chen, Ferret, Chiu, Mao-Jones, Lee, Yu, Millican, Sjoesund, Lee, Dixon, Reid, Mikuła, Wirth, Sharman, Chinaev, Thain, Bachem, Chang, Wahltinez, Bailey, Michel, Yotov, Chaabouni, Comanescu, Jana, Anil, McIlroy, Liu, Mullins, Smith, Borgeaud, Girgin, Douglas, Pandya, Shakeri, De, Klimenko, Hennigan, Feinberg, Stokowiec, hui Chen, Ahmed, Gong, Warkentin, Peran, Giang, Farabet, Vinyals, Dean, Kavukcuoglu, Hassabis, Ghahramani, Eck, Barral, Pereira, Collins, Joulin, Fiedel, Senter, Andreev, and Kenealy]{gemmateam2024gemmaopenmodelsbased-long}
Gemma Team, Thomas Mesnard, Cassidy Hardin, Robert Dadashi, Surya Bhupatiraju, Shreya Pathak, Laurent Sifre, Morgane Rivière, Mihir~Sanjay Kale, Juliette Love, Pouya Tafti, Léonard Hussenot, Pier~Giuseppe Sessa, Aakanksha Chowdhery, Adam Roberts, Aditya Barua, Alex Botev, Alex Castro-Ros, Ambrose Slone, Amélie Héliou, Andrea Tacchetti, Anna Bulanova, Antonia Paterson, Beth Tsai, Bobak Shahriari, Charline~Le Lan, Christopher~A. Choquette-Choo, Clément Crepy, Daniel Cer, Daphne Ippolito, David Reid, Elena Buchatskaya, Eric Ni, Eric Noland, Geng Yan, George Tucker, George-Christian Muraru, Grigory Rozhdestvenskiy, Henryk Michalewski, Ian Tenney, Ivan Grishchenko, Jacob Austin, James Keeling, Jane Labanowski, Jean-Baptiste Lespiau, Jeff Stanway, Jenny Brennan, Jeremy Chen, Johan Ferret, Justin Chiu, Justin Mao-Jones, Katherine Lee, Kathy Yu, Katie Millican, Lars~Lowe Sjoesund, Lisa Lee, Lucas Dixon, Machel Reid, Maciej Mikuła, Mateo Wirth, Michael Sharman, Nikolai Chinaev, Nithum Thain, Olivier Bachem,
  Oscar Chang, Oscar Wahltinez, Paige Bailey, Paul Michel, Petko Yotov, Rahma Chaabouni, Ramona Comanescu, Reena Jana, Rohan Anil, Ross McIlroy, Ruibo Liu, Ryan Mullins, Samuel~L Smith, Sebastian Borgeaud, Sertan Girgin, Sholto Douglas, Shree Pandya, Siamak Shakeri, Soham De, Ted Klimenko, Tom Hennigan, Vlad Feinberg, Wojciech Stokowiec, Yu~hui Chen, Zafarali Ahmed, Zhitao Gong, Tris Warkentin, Ludovic Peran, Minh Giang, Clément Farabet, Oriol Vinyals, Jeff Dean, Koray Kavukcuoglu, Demis Hassabis, Zoubin Ghahramani, Douglas Eck, Joelle Barral, Fernando Pereira, Eli Collins, Armand Joulin, Noah Fiedel, Evan Senter, Alek Andreev, and Kathleen Kenealy.
\newblock Gemma: Open models based on gemini research and technology, 2024.
\newblock URL \url{https://arxiv.org/abs/2403.08295}.

\bibitem[Vaswani et~al.(2017)Vaswani, Shazeer, Parmar, Uszkoreit, Jones, Gomez, Kaiser, and Polosukhin]{vaswani2017attention}
Ashish Vaswani, Noam Shazeer, Niki Parmar, Jakob Uszkoreit, Llion Jones, Aidan~N Gomez, {\L}ukasz Kaiser, and Illia Polosukhin.
\newblock Attention is all you need.
\newblock \emph{Advances in neural information processing systems}, 30, 2017.

\bibitem[Veli{\v{c}}kovi{\'c} et~al.(2017)Veli{\v{c}}kovi{\'c}, Cucurull, Casanova, Romero, Lio, and Bengio]{velivckovic2017graph-GAT}
Petar Veli{\v{c}}kovi{\'c}, Guillem Cucurull, Arantxa Casanova, Adriana Romero, Pietro Lio, and Yoshua Bengio.
\newblock Graph attention networks.
\newblock \emph{arXiv preprint arXiv:1710.10903}, 2017.

\bibitem[Wang et~al.(2024)Wang, Li, Chen, Cai, Zhu, Lin, Cao, Kong, Liu, Liu, and Sui]{wang-etal-2024-large-language-models-fair-judge-order-bias}
Peiyi Wang, Lei Li, Liang Chen, Zefan Cai, Dawei Zhu, Binghuai Lin, Yunbo Cao, Lingpeng Kong, Qi~Liu, Tianyu Liu, and Zhifang Sui.
\newblock Large language models are not fair evaluators.
\newblock In Lun-Wei Ku, Andre Martins, and Vivek Srikumar, editors, \emph{Proceedings of the 62nd Annual Meeting of the Association for Computational Linguistics (Volume 1: Long Papers)}, pages 9440--9450, Bangkok, Thailand, August 2024. Association for Computational Linguistics.
\newblock \doi{10.18653/v1/2024.acl-long.511}.
\newblock URL \url{https://aclanthology.org/2024.acl-long.511/}.

\bibitem[Wolf et~al.(2019)Wolf, Debut, Sanh, Chaumond, Delangue, Moi, Cistac, Rault, Louf, Funtowicz, et~al.]{wolf2019huggingface}
Thomas Wolf, Lysandre Debut, Victor Sanh, Julien Chaumond, Clement Delangue, Anthony Moi, Pierric Cistac, Tim Rault, R{\'e}mi Louf, Morgan Funtowicz, et~al.
\newblock Huggingface's transformers: State-of-the-art natural language processing.
\newblock \emph{arXiv preprint arXiv:1910.03771}, 2019.

\bibitem[Xu et~al.(2018)Xu, Hu, Leskovec, and Jegelka]{xu2018powerful-GIN}
Keyulu Xu, Weihua Hu, Jure Leskovec, and Stefanie Jegelka.
\newblock How powerful are graph neural networks?
\newblock \emph{arXiv preprint arXiv:1810.00826}, 2018.

\bibitem[Yang et~al.(2021)Yang, Liu, Xiao, Li, Lian, Agrawal, Singh, Sun, and Xie]{yang2021graphformers}
Junhan Yang, Zheng Liu, Shitao Xiao, Chaozhuo Li, Defu Lian, Sanjay Agrawal, Amit Singh, Guangzhong Sun, and Xing Xie.
\newblock Graphformers: Gnn-nested transformers for representation learning on textual graph.
\newblock \emph{Advances in Neural Information Processing Systems}, 34:\penalty0 28798--28810, 2021.

\bibitem[Yoon et~al.(2023)Yoon, Koh, Hooi, and Salakhutdinov]{yoon2023multimodal-graph-learning}
Minji Yoon, Jing~Yu Koh, Bryan Hooi, and Russ Salakhutdinov.
\newblock Multimodal graph learning for generative tasks.
\newblock In \emph{NeurIPS 2023 Workshop: New Frontiers in Graph Learning}, 2023.

\bibitem[Zaheer et~al.(2017)Zaheer, Kottur, Ravanbakhsh, Poczos, Salakhutdinov, and Smola]{zaheer2017deepsets}
Manzil Zaheer, Satwik Kottur, Siamak Ravanbakhsh, Barnabas Poczos, Russ~R Salakhutdinov, and Alexander~J Smola.
\newblock Deep sets.
\newblock \emph{Advances in neural information processing systems}, 30, 2017.

\bibitem[Zhang et~al.(2023)Zhang, Dong, Li, Zhang, Sun, Wang, Li, Hu, Zhang, Wu, et~al.]{zhang2023instructiontuningsurvey}
Shengyu Zhang, Linfeng Dong, Xiaoya Li, Sen Zhang, Xiaofei Sun, Shuhe Wang, Jiwei Li, Runyi Hu, Tianwei Zhang, Fei Wu, et~al.
\newblock Instruction tuning for large language models: A survey.
\newblock \emph{arXiv preprint arXiv:2308.10792}, 2023.

\bibitem[Zhang et~al.(1988)Zhang, Tanida, Itoh, and Ichioka]{zhang1988shift-invariant-CNN}
Wei Zhang, Jun Tanida, Kazuyoshi Itoh, and Yoshiki Ichioka.
\newblock Shift-invariant pattern recognition neural network and its optical architecture.
\newblock In \emph{Proceedings of annual conference of the Japan Society of Applied Physics}, volume 564. Montreal, CA, 1988.

\bibitem[Zheng et~al.(2023)Zheng, Chiang, Sheng, Zhuang, Wu, Zhuang, Lin, Li, Li, Xing, et~al.]{zheng2023judgingLLM-as-a-Judge-MT-Arena}
Lianmin Zheng, Wei-Lin Chiang, Ying Sheng, Siyuan Zhuang, Zhanghao Wu, Yonghao Zhuang, Zi~Lin, Zhuohan Li, Dacheng Li, Eric Xing, et~al.
\newblock Judging llm-as-a-judge with mt-bench and chatbot arena.
\newblock \emph{Advances in Neural Information Processing Systems}, 36:\penalty0 46595--46623, 2023.

\bibitem[Zheng et~al.(2025)Zheng, Gan, Chen, Qi, Liang, and Yu]{zheng2025llminmedicinesurvey}
Yanxin Zheng, Wensheng Gan, Zefeng Chen, Zhenlian Qi, Qian Liang, and Philip~S Yu.
\newblock Large language models for medicine: a survey.
\newblock \emph{International Journal of Machine Learning and Cybernetics}, 16\penalty0 (2):\penalty0 1015--1040, 2025.

\bibitem[Zhu et~al.(2023)Zhu, Wang, and Wang]{zhu2023judgelm}
Lianghui Zhu, Xinggang Wang, and Xinlong Wang.
\newblock Judgelm: Fine-tuned large language models are scalable judges.
\newblock \emph{arXiv preprint arXiv:2310.17631}, 2023.

\bibitem[Zhuang et~al.(2024)Zhuang, Zhuang, Koopman, and Zuccon]{zhuang2024setwise-prompting}
Shengyao Zhuang, Honglei Zhuang, Bevan Koopman, and Guido Zuccon.
\newblock A setwise approach for effective and highly efficient zero-shot ranking with large language models.
\newblock In \emph{Proceedings of the 47th International ACM SIGIR Conference on Research and Development in Information Retrieval}, pages 38--47, 2024.

\bibitem[Zong et~al.(2024)Zong, Yu, Chavhan, Zhao, and Hospedales]{zongfool}
Yongshuo Zong, Tingyang Yu, Ruchika Chavhan, Bingchen Zhao, and Timothy Hospedales.
\newblock Fool your (vision and) language model with embarrassingly simple permutations.
\newblock In \emph{Forty-first International Conference on Machine Learning}, 2024.

\end{thebibliography}
\bibliographystyle{plainnat}

\makeatletter
\if@preprint

\else
    empty

\fi

\clearpage
\appendix

\section{Impact}
\label{app:impact}

We do not foresee any direct negative impacts from this work. On the contrary, we believe this work, as with any other work on model robustness, can contribute positively to LLM applications, especially in high-risk scenarios. Moreover, this work can be used to develop more robust evaluation approaches, which can help move the wider field forward.

\section{SetPE algorithm}
\label{app:setPE_alg}

\begin{algorithm}
\caption{Python-like pseudocode for calculating Set Position Encoding (SetPE) positions.}
\label{alg:setPE}
\small
\medskip
{\tt
\color{teal}
\# input - mixed list of token ids and sets of token id lists \\
\# vocab - token vocabulary
}

\medskip
{\tt
\color{black}
pos = [ ] \textcolor{teal}{\# list of SetPE positions}\\
ind = 0 \textcolor{teal}{\# running position index}\\
for q in input: \textcolor{teal}{\# iterate through elements of the input}\\
\verb|  |if q in vocab: \textcolor{teal}{\# if element q is a token}\\
\verb|    |pos.append(ind)\\
\verb|    |ind += 1\\
\verb|  |else: \textcolor{teal}{\# if element q is a set of token lists}\\
\verb|    |for s in q: \textcolor{teal}{\# iterate through set of token lists}\\
\verb|      |\textcolor{teal}{\# append consecutive positions for each token in s:}\\
\verb|      |pos = pos + list(range(ind, ind+len(s)))\\
\verb|    |total\_tokens\_q = sum([len(s) for s in q])\\
\verb|    |ind += total\_tokens\_q\\
return pos
}
\medskip
\end{algorithm}

\section{Proofs}
\label{app:proofs}
\printProofs

\section{Additional experimental details}
\label{app:add-experimental-details}

\subsection{Instruction finetuning}
\label{app:instruction-finetuning}

Finally, we introduce \emph{Instruction Finetuning} as the proposed method relies on this training approach.
Many LLMs use some form of instruction tuning during the (pre-)training process \citep{ouyang2022instructiontuning, zhang2023instructiontuningsurvey}. 
It involves a dataset $\mathcal{D}$ of instruction-answer (or prompt-response) pairs $\mathcal{D} = \left\{(q,a)\right\}_{i=1}^N$. The training objective is to maximize the probability of autoregressively generating $a$ given $q$, i.e., to maximize
\begin{equation}
    p(a \mid q) = \prod_{i=1, \ldots, \abs{a}} p(a_i \mid q, a_1, \ldots, a_{i-1}).
\end{equation}

We use instruction tuning for training all our models to align to the new position encoding and attention masking setups. Since all probabilities in the product are conditional on $q$, $q$ can be encoded without a causal mask, i.e., tokens in $q$ could attend to earlier tokens in $q$.
This makes instruction tuning ideal for our scenario.

\subsection{Dataset details}
\label{app:dataset-metadata}

We get all the datasets from HuggingFace Datasets \citep{lhoest-etal-2021-datasets-huggingface}. \cref{tab:dataset-metadata} provides metadata and \cref{tab:dataset-licenses} provides licensing details for each dataset.

\begin{table}[H]
    \centering
    \setlength{\tabcolsep}{4.4pt}
    \caption{HuggingFace Datasets path, number of train/evaluation samples, number of choices per question, and number of answer choice permutations for each dataset in this paper.}
    \begin{tabular}{llcccc}
        \toprule
        Dataset & Path & \# Train. Samples & \# Eval. Samples & $k$ & $k!$ \\
        \midrule
        UltraFeedback \citep{cui2024ultrafeedbackboostinglanguagemodels,kopiczko2024bitune}
        & openbmb/UltraFeedback & & - & - & - \\
        PIQA \citep{bisk2019piqa-PIQA}
        & piqa & 16113 & 3084 & 2 & 2 \\
        ARC \citep{allenai:arc-ARC}
        & allenai/ai2\_arc & 1119 & 1172 & 4 & 24 \\
        CSQA \citep{talmor-etal-2019-commonsenseqa-CSQA}
        & tau/commonsense\_qa & 9741 & 1140 & 5 & 120 \\
        SIQA \citep{sap-etal-2019-social-SIQA}
        & social\_i\_qa & 33410 & 1954 & 3 & 6 \\
        \bottomrule
    \end{tabular}
    \label{tab:dataset-metadata}
\end{table}

\begin{table}[H]
    \centering
    \setlength{\tabcolsep}{4.4pt}
    \caption{Dataset Licenses.}
    \begin{tabular}{ll}
        \toprule
        Dataset & License \\
        \midrule
        UltraFeedback \citep{cui2024ultrafeedbackboostinglanguagemodels,kopiczko2024bitune}
        & MIT license \\
        PIQA \citep{bisk2019piqa-PIQA}
        & unknown \\
        ARC \citep{allenai:arc-ARC}
        & CC-BY-SA 4.0 \\
        CSQA \citep{talmor-etal-2019-commonsenseqa-CSQA}
        & MIT license \\
        SIQA \citep{sap-etal-2019-social-SIQA}
        & unknown \\
        \bottomrule
    \end{tabular}
    \label{tab:dataset-licenses}
\end{table}

\clearpage

\subsection{Prompt templates}
\label{app:prompts}

\paragraph{Original templates.}We provide all the original templates used to create prompt-response pairs for each dataset, for both training and evaluation.

\begin{table}[H]
    \centering
    \setlength{\tabcolsep}{6pt}
    \caption{Original prompt templates for each dataset.}
    \begin{longtable}{@{}p{.20\textwidth}p{.45\textwidth}p{.27\textwidth}@{}}
    \toprule
        Dataset & Original Prompt 
        & Response \\
        \midrule
        UltraFeedback \citep{cui2024ultrafeedbackboostinglanguagemodels}
        & \makecell[tl]{Question: \{instruction\} \\ \\Answer:}
        & \makecell[tl]{\{answer\}<EOS>} \\
        \midrule
        PIQA \citep{bisk2019piqa-PIQA}
        & \makecell[tl]{Question: \{question\} \\ \\Answer:}
        & \makecell[tl]{\{answer\}<EOS>} \\
        \midrule
        ARC \citep{allenai:arc-ARC}
        & \makecell[tl]{Question: \{question\} \\ \\Answer:}
        & \makecell[tl]{\{answer\}<EOS>} \\
        \midrule
        CSQA \citep{talmor-etal-2019-commonsenseqa-CSQA}
        & \makecell[tl]{Question: \{question\} \\ \\Answer:}
        & \makecell[tl]{\{answer\}<EOS>} \\
        \midrule
        SIQA \citep{sap-etal-2019-social-SIQA}
        & \makecell[tl]{
        Question: Given the context, answer \\correctly the question. \\
        Context: \{context\} \\ 
        Question: \{question\} \\ \\
        Choices: \\
        (0) \{choice0\} \\
        (1) \{choice1\} \\
        (2) \{choice2\} \\ \\
        Answer:
        }
        & \makecell[tl]{(\{answer\_index\})<EOS>} \\
        \bottomrule
    \end{longtable}
    \label{tab:original-prompts}
\end{table}

\clearpage

\paragraph{Modified templates.}We provide all the modified templates used to create prompt-response pairs for each dataset, for both training and evaluation.

\begin{table}[H]
    \centering
    \setlength{\tabcolsep}{6pt}
    \caption{Modified prompt templates for each dataset.}
    \begin{longtable}{@{}p{.20\textwidth}p{.45\textwidth}p{.27\textwidth}@{}}
    \toprule
        Dataset 
        & Modified Prompt 
        & Response \\
        \midrule
        UltraFeedback \citep{cui2024ultrafeedbackboostinglanguagemodels}
        & \makecell[tl]{Question: \{instruction\} \\ \\Answer:}
        & \makecell[tl]{\{answer\}<EOS>} \\
        \midrule
        PIQA \citep{bisk2019piqa-PIQA}
        & \makecell[tl]{Question: \{question\} \\ \\Choices: \\\{choice0\} \\\{choice1\} \\ \\Answer:}
        & \makecell[tl]{\{answer\}<EOS>} \\
        \midrule
        ARC \citep{allenai:arc-ARC}
        & \makecell[tl]{Question: \{question\} \\ \\Choices: \\\{choice0\} \\\{choice1\} \\\{choice2\} \\\{choice3\}  \\ \\Answer:}
        & \makecell[tl]{\{answer\}<EOS>} \\
        \midrule
        CSQA \citep{talmor-etal-2019-commonsenseqa-CSQA}
        & \makecell[tl]{Question: \{question\} \\ \\Choices: \\\{choice0\} \\\{choice1\} \\\{choice2\} \\\{choice3\} \\\{choice4\}  \\ \\Answer:}
        & \makecell[tl]{\{answer\}<EOS>} \\
        \midrule
        SIQA \citep{sap-etal-2019-social-SIQA}
        & \makecell[tl]{
        Question: Given the context, answer \\correctly the question. \\
        Context: \{context\} \\ 
        Question: \{question\} \\ \\
        Choices: \\
        \{choice0\} \\
        \{choice1\} \\
        \{choice2\} \\ \\
        Answer:
        }
        & \makecell[tl]{\{answer\}<EOS>} \\
        \bottomrule
    \end{longtable}
    \label{tab:modified-prompts}
\end{table}

\clearpage

\subsection{Hyperparameter settings}
\label{app:hyperparameters}

\begin{table}[H]
    \centering
    \caption{Hyperparameters shared across models and datasets.}
    \begin{tabular}{lc}
        \toprule
        Hyperparameter & Value \\
        \midrule
        GPUs & 1 \\
        Optimizer & AdamW \\
        LR Scheduler & Linear \\
        Weight Decay & 0.0 \\
        Batch Size & 10 \\
        Accumulation Steps & 10 \\
        Warmup Steps & 300 (or 10\% of update steps) \\
        Update Steps & 3000 \\
        Random Seed & 42 \\
        \bottomrule
    \end{tabular}
    \label{tab:general-hyperparameters}
\end{table}

We tune the learning rate for each model and dataset using a logarithmic scale: [1e-4, 3e-4, 1e-3, 3e-3]. The final (best) learning rates are presented in \cref{tab:learning-rates}.

\begin{table}[H]
    \centering
    \caption{Learning rates for all final baseline and Set-LLM models on all datasets.}
    \begin{tabular}{llccccc}
        \toprule
        LLM & Model & Ultra. & PIQA & ARC & CSQA & SIQA \\
        \midrule
        \multirow{2}{*}{Gemma 2B} 
        & Causal Mask+PE$^{\text{Ultra}}$ & 3e-4 & 1e-3 & 1e-3 & 1e-3 & 1e-3 \\
        & SetMask+SetPE$^{\text{Ultra}}$ & 3e-4 & 1e-3 & 1e-3 & 1e-3 & 1e-3 \\
        \midrule
        \multirow{2}{*}{Gemma 7B} 
        & Causal Mask+PE$^{\text{Ultra}}$ & 3e-4 & 3e-4 & 3e-4 & 3e-4 & 3e-4 \\
        & SetMask+SetPE$^{\text{Ultra}}$ & 3e-4 & 3e-4 & 3e-4 & 3e-4 & 3e-4 \\
        \midrule
        \multirow{2}{*}{Llama 3.2 1B} 
        & Causal Mask+PE$^{\text{Ultra}}$ & 1e-3 & 1e-3 & 1e-3 & 1e-3 & 1e-3 \\
        & SetMask+SetPE$^{\text{Ultra}}$ & 1e-3 & 1e-3 & 1e-3 & 1e-3 & 1e-3 \\
        \midrule
        \multirow{2}{*}{Llama 3.2 3B} 
        & Causal Mask+PE$^{\text{Ultra}}$ & 3e-4 & 1e-3 & 1e-3 & 1e-3 & 1e-3 \\
        & SetMask+SetPE$^{\text{Ultra}}$ & 3e-4 & 1e-3 & 1e-3 & 1e-3 & 1e-3 \\
        \midrule
        \multirow{2}{*}{Llama 3.1 8B} 
        & Causal Mask+PE$^{\text{Ultra}}$ & 3e-4 & 3e-4 & 3e-4 & 3e-4 & 3e-4 \\
        & SetMask+SetPE$^{\text{Ultra}}$ & 1e-3 & 3e-4 & 3e-4 & 3e-4 & 3e-4 \\
        \bottomrule
    \end{tabular}
    \label{tab:learning-rates}
\end{table}

\begin{table}[H]
    \centering
    \caption{We use LoRA \citep{hu2022lora} to train all our models. We use the HuggingFace PEFT library \citep{2022peft-huggingface} with default hyperparameter values, unless listed.}
    \begin{tabular}{lc}
        \toprule
        Hyperparameter & Value \\
        \midrule
        Rank & 8 \\
        Alpha & 1 \\
        Target Modules & All linear layers of MLP and Self-Attention \\
        \bottomrule
    \end{tabular}
    \label{tab:peft-hyperparameters}
\end{table}

\section{Additional experimental results}

\subsection{Additional pretraining}
\label{app:pre-finetuning}

To help the models adapt to the architectural changes, we experiment with additional pretraining.
We use a high-quality subset (approximately 10k examples) of the cleaned UltraFeedback dataset \citep{cui2024ultrafeedbackboostinglanguagemodels}, attained by following the data preprocessing steps in \citep{kopiczko2024bitune}. 
This additional pretraining aims to help the new, adapted model architectures better adapt to the architectural changes. We therefore hypothesize that adapted models will benefit more from this data than the unaltered baseline models.

We present complete results with and without additional pretraining in \cref{tab:pretraining}. Consistent with our hypothesis, the additional data significantly improves the performance of Prefix Mask+PE, Prefix Mask+SetPE, and SetMask+SetPE, especially on ARC-Challenge, but does not improve the performance of the finetuned Causal Mask+PE. However, the pretrained Causal Mask+PE results improve the most, suggesting that the pretrained base model is not particularly well-suited to the benchmark task setups.

\begin{table}[ht]
    \centering
    \setlength{\tabcolsep}{6pt}
    \small
    \caption{Results of (pre-)finetuning Gemma 2B on UltraFeedback with the respective attention mask and positional encoding. 
    The $^{*}$ indicates results using the original dataset prompts, which for the PIQA, ARC, and CSQA benchmark only contain the question. All other results use modified prompts, where the choices are provided with the question. 
    $^{\dagger}$The CSQA dataset has exactly $5$ choices for each question, but we run the adversarial search for only $24$ permutations.
    All scores are accuracies $(\%)$.
    }
    \begin{tabular}{llc @{\hspace{1\tabcolsep}} cc @{\hspace{1\tabcolsep}} cc @{\hspace{1\tabcolsep}} cc @{\hspace{1\tabcolsep}} c}
         \toprule
         Model & Training & \multicolumn{2}{c}{PIQA} & \multicolumn{2}{c}{ARC} & \multicolumn{2}{c}{CSQA} & \multicolumn{2}{c}{SIQA}\\
         & & Std. & Adv. & Std. & Adv. & Std. & Adv.$^{\dagger}$ & Std. & Adv.  \\
         \midrule
         Causal Mask+PE$^{*}$ & Pretrained
         & 76.77 &    & 37.80 & 
         & 51.76 &    & 37.26 &  \\           
         Causal Mask+PE$^{*}$ & Finetuned
         & 79.82 &    & 45.39 & 
         & 68.80 &    & 75.95 &  \\          
         \midrule
         Causal Mask+PE & Pretrained
         & 57.45 & \textcolor{blue}{30.96}   & 36.03 & \textcolor{blue}{\phantom{1}7.68}
         & 34.92 & \textcolor{blue}{16.46}   & 39.29 & \textcolor{blue}{12.74} \\
         Causal Mask+PE$^{\text{Ultra}}$ & Pretrained
         & 68.31 & \textcolor{blue}{50.49}   & 43.18 & \textcolor{blue}{14.93}
         & 45.21 & \textcolor{blue}{27.35}   & 46.72 & \textcolor{blue}{16.84} \\
         Causal Mask+PE & Finetuned
         & 84.11 & \textcolor{blue}{76.77}   & 55.20 & \textcolor{blue}{23.72}
         & 78.31 & \textcolor{blue}{69.62}   & 74.80 & \textcolor{blue}{63.00} \\
         Causal Mask+PE$^{\text{Ultra}}$ & Finetuned
         & 83.98 & \textcolor{blue}{77.31}   & 56.32 & \textcolor{blue}{26.88}
         & 77.89 & \textcolor{blue}{68.47}   & 74.33 & \textcolor{blue}{63.97}\\
         Prefix Mask+PE & Finetuned
         & 82.78 & \textcolor{blue}{76.50}   & 57.62 & \textcolor{blue}{27.47}
         & 78.98 & \textcolor{blue}{71.01}   & 74.36 & \textcolor{blue}{65.66} \\
         Prefix Mask+PE$^{\text{Ultra}}$ & Finetuned
         & 84.93 & \textcolor{blue}{79.11}   & 61.26 & \textcolor{blue}{34.47}
         & 79.28 & \textcolor{blue}{70.19}   & 75.18 & \textcolor{blue}{67.45} \\
         Prefix Mask+SetPE & Finetuned  
         & 81.23 & \textcolor{blue}{81.23}   & 51.28 & \textcolor{blue}{51.28}   
         & 77.31 & \textcolor{blue}{77.31}   & 71.24 & \textcolor{blue}{71.24} \\
         Prefix Mask+SetPE$^{\text{Ultra}}$ & Finetuned
         & 81.88 & \textcolor{blue}{81.88}   & 56.48 & \textcolor{blue}{56.48}
         & 77.97 & \textcolor{blue}{77.97}   & 73.54 & \textcolor{blue}{73.54} \\
         SetMask+SetPE & Finetuned
         & 84.33 & \textcolor{blue}{84.33}   & 57.76 & \textcolor{blue}{57.76}
         & 79.93 & \textcolor{blue}{79.93}   & 75.38 & \textcolor{blue}{75.38} \\
         SetMask+SetPE$^{\text{Ultra}}$ & Finetuned
         & \textbf{85.80} & \textbf{\textcolor{blue}{85.80}}   & \textbf{65.02} & \textbf{\textcolor{blue}{65.02}}
         & \textbf{80.18} & \textbf{\textcolor{blue}{80.18}}   & \textbf{76.15} & \textbf{\textcolor{blue}{76.15}} \\
         \bottomrule
         \vspace{-0.25cm} \\
         \multicolumn{10}{l}{$^{*}$Results with original prompts 
         \quad $^{\text{Ultra}}$Additional pretraining
         \quad $^{\dagger}$Only first 24 (of 120) permutations tested
         }
    \end{tabular}
    \label{tab:pretraining}
\end{table}

\subsection{Majority vote}
\label{app:majority_vote}

Additional results with majority vote are presented in \cref{tab:majority_vote_ft}

\begin{table}[ht]
    \centering
    \setlength{\tabcolsep}{3.4pt}
    \small
    \caption{Majority Vote results with Gemma 2B as the base model. 
    The $^{*}$ indicates results using the original dataset prompts, which for the PIQA, ARC, and CSQA benchmark only contain the question. All other results use modified prompts, where the choices are provided with the question. 
    $^{\dagger}$The CSQA dataset has exactly $5$ choices for each question, but we run the adversarial search for only $24$ permutations.
    All scores are accuracies $(\%)$.
    }
    \begin{tabular}{lllc @{\hspace{1\tabcolsep}} cc @{\hspace{1\tabcolsep}} cc @{\hspace{1\tabcolsep}} cc @{\hspace{1\tabcolsep}} c}
         \toprule
         Model & Training & Eval. Mode & \multicolumn{2}{c}{PIQA} & \multicolumn{2}{c}{ARC} & \multicolumn{2}{c}{CSQA} & \multicolumn{2}{c}{SIQA}\\
         & & & Std. & Adv. & Std. & Adv. & Std. & Adv.$^{\dagger}$ & Std. & Adv.  \\
         \midrule
         Causal Mask+PE$^{*}$ & Pretrained & Single run
         & 76.77 &    & 37.80 & 
         & 51.76 &    & 37.26 & \\
         Causal Mask+PE$^{*}$ & Finetuned & Single run
         & 79.82 &    & 45.39 & 
         & 68.80 &    & 75.95 & \\
         \midrule
         Causal Mask+PE$^{\text{Ultra}}$ & Pretrained & Single run
         & 68.31 & \textcolor{blue}{50.49}   & 43.18 & \textcolor{blue}{14.93}
         & 45.21 & \textcolor{blue}{27.35}   & 46.72 & \textcolor{blue}{16.84} \\
         Causal Mask+PE$^{\text{Ultra}}$ & Pretrained & Majority Vote
         & 68.12 & \textcolor{blue}{68.12}   & 46.16 & \textcolor{blue}{46.16}
         & 45.95 & \textcolor{blue}{45.95}   & 48.62 & \textcolor{blue}{48.62} \\
         Causal Mask+PE$^{\text{Ultra}}$ & Finetuned & Single run
         & 83.98 & \textcolor{blue}{77.31}   & 56.32 & \textcolor{blue}{26.88}
         & 77.89 & \textcolor{blue}{68.47}   & 74.33 & \textcolor{blue}{63.97}\\
         Causal Mask+PE$^{\text{Ultra}}$ & Finetuned & Majority Vote
         & 83.57 & \textcolor{blue}{83.57}   & 59.56 & \textcolor{blue}{59.56}
         & 78.46 & \textcolor{blue}{78.46}   & 75.23 & \textcolor{blue}{75.23} \\
         Causal Mask+NoPE$^{\text{Ultra}}$ & Finetuned & Single run
         & 75.03 & \textcolor{blue}{64.53}   & 37.01 & \textcolor{blue}{14.68}
         & 68.87 & \textcolor{blue}{57.41}   & 65.57 & \textcolor{blue}{51.79} \\
         Causal Mask+NoPE$^{\text{Ultra}}$ & Finetuned & Majority Vote
         & 75.35 & \textcolor{blue}{75.35}   & 38.82 & \textcolor{blue}{38.82}
         & 69.45 & \textcolor{blue}{69.45}   & 66.63 & \textcolor{blue}{66.63} \\
         Prefix Mask+PE$^{\text{Ultra}}$ & Finetuned & Single run
         & 84.93 & \textcolor{blue}{79.11}   & 61.26 & \textcolor{blue}{34.47}
         & 79.28 & \textcolor{blue}{70.19}   & 75.18 & \textcolor{blue}{67.45} \\
         Prefix Mask+PE$^{\text{Ultra}}$ & Finetuned & Majority Vote
         & 85.36 & \textcolor{blue}{85.36}   & 64.08 & \textcolor{blue}{64.08}
         & 79.77 & \textcolor{blue}{79.77}   & 75.84 & \textcolor{blue}{75.84} \\
         SetMask+SetPE$^{\text{Ultra}}$ & Finetuned & Single run
         & \textbf{85.80} & \textbf{\textcolor{blue}{85.80}}   & \textbf{65.02} & \textbf{\textcolor{blue}{65.02}}
         & \textbf{80.18} & \textbf{\textcolor{blue}{80.18}}   & \textbf{76.15} & \textbf{\textcolor{blue}{76.15}} \\
         \bottomrule
         \vspace{-0.25cm} \\
         \multicolumn{11}{l}{$^{*}$Results with original prompts 
         \quad $^{\text{Ultra}}$Additional pretraining
         \quad $^{\dagger}$Only first 24 (of 120) permutations tested
         }
    \end{tabular}
    \label{tab:majority_vote_ft}
\end{table}

\clearpage

\subsection{Different base LLMs}
\label{app:all-base-LLMs}

Additional results with different base LLMs are presented in \cref{tab:all_models_full}.

\begin{table}[ht]
    \centering
    \setlength{\tabcolsep}{3pt}
    \small
    \caption{Performance with different base LLMs. All results (except Pretrained$^*$) are (pre-)finetuned on the ultrafeedback dataset.
    $^{\dagger}$The CSQA dataset has exactly $5$ choices for each question, but we run the adversarial search for only $24$ permutations.
    All scores are accuracies $(\%)$.
    }
    \begin{tabular}{lllc @{\hspace{1\tabcolsep}} cc @{\hspace{1\tabcolsep}} cc @{\hspace{1\tabcolsep}} cc @{\hspace{1\tabcolsep}} c}
         \toprule
         LLM & Model & Training & \multicolumn{2}{c}{PIQA} & \multicolumn{2}{c}{ARC} & \multicolumn{2}{c}{CSQA} & \multicolumn{2}{c}{SIQA}\\
         & & & Rand. & Adv. & Rand. & Adv. & Rand. & Adv.$^{\dagger}$ & Rand. & Adv. \\
         \midrule
         \multirow{5}{*}{Gemma 2B}
         & Causal Mask+PE$^{*}$ & Pretrained
         & 76.77 &    & 37.80 & 
         & 51.76 &    & 37.26 &  \\
         & Causal Mask+PE$^{\text{Ultra}}$ & Pretrained
         & 68.31 & \textcolor{blue}{50.49}   & 43.18 & \textcolor{blue}{14.93}
         & 45.21 & \textcolor{blue}{27.35}   & 46.72 & \textcolor{blue}{16.84} \\
         & Causal Mask+PE$^{\text{Ultra}}$ & Finetuned
         & 83.98 & \textcolor{blue}{77.31}   & 56.32 & \textcolor{blue}{26.88}
         & 77.89 & \textcolor{blue}{68.47}   & 74.33 & \textcolor{blue}{63.97}\\
         & \quad + Majority Vote & Finetuned
         & 84.17 & \textcolor{blue}{84.17}   & 60.15 & \textcolor{blue}{60.15}
         & 78.71 & \textcolor{blue}{78.71}   & 75.38 & \textcolor{blue}{75.38} \\
         & SetMask+SetPE$^{\text{Ultra}}$ & Finetuned
         & \textbf{85.80} & \textbf{\textcolor{blue}{85.80}}   & \textbf{65.02} & \textbf{\textcolor{blue}{65.02}}
         & \textbf{80.18} & \textbf{\textcolor{blue}{80.18}}   & \textbf{76.15} & \textbf{\textcolor{blue}{76.15}} \\
         \midrule
         \multirow{5}{*}{Gemma 7B}
         & Causal Mask+PE$^{*}$ & Pretrained
         & 80.41 &    & 43.77 & 
         & 62.16 &    & 65.71 &  \\
         & Causal Mask+PE$^{\text{Ultra}}$ & Pretrained
         & 86.21 & \textcolor{blue}{78.67}   & 79.68 & \textcolor{blue}{56.23}
         & 69.96 & \textcolor{blue}{47.17}   & 70.34 & \textcolor{blue}{49.69} \\
         & Causal Mask+PE$^{\text{Ultra}}$ & Finetuned
         & 92.82 & \textcolor{blue}{89.45}   & 83.52 & \textcolor{blue}{64.33}
         & 85.45 & \textcolor{blue}{79.12}   & 80.93 & \textcolor{blue}{74.10} \\
         & \quad + Majority Vote & Finetuned
         & 92.66 & \textcolor{blue}{92.66}   & \textbf{85.58} & \textbf{\textcolor{blue}{85.58}}
         & \textbf{85.75} & \textbf{\textcolor{blue}{85.75}}   & 81.10 & \textcolor{blue}{81.10} \\
         & SetMask+SetPE$^{\text{Ultra}}$ & Finetuned
         & \textbf{92.98} & \textbf{\textcolor{blue}{92.98}}   & 83.45 & \textcolor{blue}{83.45}
         & 84.93 & \textcolor{blue}{84.93}   & \textbf{81.12} & \textbf{\textcolor{blue}{81.12}} \\
         \midrule
         \multirow{5}{*}{Llama 3.2 1B}
         & Causal Mask+PE$^{*}$ & Pretrained
         & 74.32 &    & 35.41 & 
         & 55.77 &    & 51.59 & \\
         & Causal Mask+PE$^{\text{Ultra}}$ & Pretrained
         & 63.03 & \textcolor{blue}{40.48}   & 40.12 & \textcolor{blue}{\phantom{1}9.13}
         & 45.29 & \textcolor{blue}{21.21}   & 49.27 & \textcolor{blue}{21.39} \\
         & Causal Mask+PE$^{\text{Ultra}}$ & Finetuned
         & 79.57 & \textcolor{blue}{71.33}   & 53.61 & \textcolor{blue}{21.93}
         & 74.50 & \textcolor{blue}{64.21}   & 71.84 & \textcolor{blue}{62.79} \\
         & \quad + Majority Vote & Finetuned
         & 79.49 & \textcolor{blue}{79.49}   & 57.17 & \textcolor{blue}{57.17}
         & 75.51 & \textcolor{blue}{75.51}   & 71.85 & \textcolor{blue}{71.85} \\
         & SetMask+SetPE$^{\text{Ultra}}$ & Finetuned
         & \textbf{81.66} & \textbf{\textcolor{blue}{81.66}}   & \textbf{59.30} & \textbf{\textcolor{blue}{59.30}}
         & \textbf{76.66} & \textbf{\textcolor{blue}{76.66}}   & \textbf{72.47} & \textbf{\textcolor{blue}{72.47}} \\
         \midrule
         \multirow{5}{*}{Llama 3.2 3B}
         & Causal Mask+PE$^{*}$ & Pretrained
         & 76.33 &    & 43.94 & 
         & 61.92 &    & 65.97 & \\
         & Causal Mask+PE$^{\text{Ultra}}$ & Pretrained
         & 76.41 & \textcolor{blue}{64.09}   & 68.83 & \textcolor{blue}{39.93}
         & 66.85 & \textcolor{blue}{44.31}   & 66.56 & \textcolor{blue}{47.80} \\
         & Causal Mask+PE$^{\text{Ultra}}$ & Finetuned
         & 86.92 & \textcolor{blue}{81.72}   & 74.16 & \textcolor{blue}{53.07}
         & 81.32 & \textcolor{blue}{74.94}   & 77.54 & \textcolor{blue}{70.42} \\
         & \quad + Majority Vote & Finetuned
         & 86.83 & \textcolor{blue}{86.83}   & \textbf{76.37} & \textbf{\textcolor{blue}{76.37}}
         & 81.57 & \textcolor{blue}{81.57}   & 77.99 & \textcolor{blue}{77.99} \\
         & SetMask+SetPE$^{\text{Ultra}}$ & Finetuned
         & \textbf{88.41} & \textbf{\textcolor{blue}{88.41}}   & 75.85 & \textcolor{blue}{75.85}
         & \textbf{83.29} & \textbf{\textcolor{blue}{83.29}}   & \textbf{80.30} & \textbf{\textcolor{blue}{80.30}} \\
         \midrule
         \multirow{5}{*}{Llama 3.1 8B}
         & Causal Mask+PE$^{*}$ & Pretrained
         & 80.09 &    & 53.41 &  
         & 66.50 &    & 69.34 & \\
         & Causal Mask+PE$^{\text{Ultra}}$ & Pretrained
         & 83.30 & \textcolor{blue}{72.36}   & 78.75 & \textcolor{blue}{56.66}   
         & 72.67 & \textcolor{blue}{53.81}   & 70.91 & \textcolor{blue}{54.96} \\
         & Causal Mask+PE$^{\text{Ultra}}$ & Finetuned
         & 90.81 & \textcolor{blue}{86.29}   & 83.04 & \textcolor{blue}{64.51}   
         & 83.96 & \textcolor{blue}{77.89}   & 80.77 & \textcolor{blue}{73.90} \\
         & \quad + Majority Vote & Finetuned
         & 90.75 & \textcolor{blue}{90.75}   & \textbf{85.32} & \textbf{\textcolor{blue}{85.32}} 
         & 84.11 & \textcolor{blue}{84.11}   & 81.12 & \textcolor{blue}{81.12} \\
         & SetMask+SetPE$^{\text{Ultra}}$ & Finetuned
         & \textbf{91.62} & \textbf{\textcolor{blue}{91.62}}   & 84.13 & \textcolor{blue}{84.13} 
         & \textbf{85.34} & \textbf{\textcolor{blue}{85.34}}   & \textbf{81.47} & \textbf{\textcolor{blue}{81.47}} \\
         \bottomrule
         \vspace{-0.25cm} \\
         \multicolumn{11}{l}{$^{*}$Results with original prompts 
         \quad $^{\text{Ultra}}$Additional pretraining
         \quad $^{\dagger}$Only first 24 (of 120) permutations tested
         }
    \end{tabular}
    \label{tab:all_models_full}
\end{table}

\subsection{Out-of-distribution performance}
\label{app:OOD}

Comparison of the out-of-distribution performance finetuned Set-LLM and baseline models. The base model is Gemma 2B.

\begin{table}[ht]
    \centering
    \setlength{\tabcolsep}{3.5pt}
    \small
    \caption{Results on out-of-distribution datasets using Gemma 2B as the base model. In-distribution results are grayed out.
    All scores are accuracies $(\%)$.
    }
    \begin{tabular}{lcc @{\hspace{1\tabcolsep}} cc @{\hspace{1\tabcolsep}} cc @{\hspace{1\tabcolsep}} cc @{\hspace{1\tabcolsep}} c}
         \toprule
         Model & Finetune & \multicolumn{2}{c}{PIQA} & \multicolumn{2}{c}{ARC} & \multicolumn{2}{c}{CSQA} & \multicolumn{2}{c}{SIQA}\\
         & Dataset & Std. & Adv. & Std. & Adv. & Std. & Adv.$^{\dagger}$ & Std. & Adv.  \\
         \midrule
         Pretrained                 
         & - & 57.45 & \textcolor{blue}{30.96}   & 36.03 & \textcolor{blue}{\phantom{1}7.68}   
         & 34.92 & \textcolor{blue}{16.46}   & 39.29 & \textcolor{blue}{12.74} \\
         Pretrained$^{\text{Ultra}}$
         & - & 68.31 & \textcolor{blue}{50.49}   & 43.18 & \textcolor{blue}{14.93}
         & 45.21 & \textcolor{blue}{27.35}   & 46.72 & \textcolor{blue}{16.84} \\
         \midrule
         Causal Mask+PE$^{\text{Ultra}}$
         &  
         & \textcolor{gray}{83.98} & \textcolor{lightgray}{77.31}   & 54.45 & \textcolor{blue}{26.79}   
         & 59.68 & \textcolor{blue}{41.52}   & 56.79 & \textcolor{blue}{35.11} \\
         \quad + Majority Vote 
         & PIQA 
         & \textcolor{gray}{83.57} & \textcolor{lightgray}{83.57}   & 56.74 & \textcolor{blue}{56.74}
         & 61.51 & \textcolor{blue}{61.51}   & \textbf{58.34} & \textbf{\textcolor{blue}{58.34}} \\
         SetMask+SetPE$^{\text{Ultra}}$
         &  
         & \textbf{\textcolor{gray}{85.80}} & \textbf{\textcolor{lightgray}{85.80}}   & \textbf{58.02} & \textbf{\textcolor{blue}{58.02}}
         & \textbf{63.47} & \textbf{\textcolor{blue}{63.47}}   & 56.86 & \textcolor{blue}{56.86} \\

         \midrule
         Causal Mask+PE$^{\text{Ultra}}$
         &  
         & 67.27 & \textcolor{blue}{47.55}   & \textcolor{gray}{56.32} & \textcolor{lightgray}{26.88}   
         & 57.61 & \textcolor{blue}{35.71}   & 54.91 & \textcolor{blue}{31.73} \\
         \quad + Majority Vote 
         & ARC 
         & 66.92 & \textcolor{blue}{66.92}   & \textcolor{gray}{59.56} & \textcolor{lightgray}{59.56}
         & 59.38 & \textcolor{blue}{59.38}   & 56.86 & \textcolor{blue}{56.86} \\
         SetMask+SetPE$^{\text{Ultra}}$
         &  
         & \textbf{68.61} & \textbf{\textcolor{blue}{68.61}}   & \textbf{\textcolor{gray}{65.02}} & \textbf{\textcolor{lightgray}{65.02}}
         & \textbf{63.39} & \textbf{\textcolor{blue}{63.39}}   & \textbf{60.64} & \textbf{\textcolor{blue}{60.64}} \\

         \midrule
         Causal Mask+PE$^{\text{Ultra}}$
         &  
         & 71.84 & \textcolor{blue}{56.64}   & 51.63 & \textcolor{blue}{26.45}   
         & \textcolor{gray}{77.89} & \textcolor{lightgray}{68.47}   & 55.72 & \textcolor{blue}{43.14} \\
         \quad + Majority Vote 
         & CSQA 
         & \textbf{72.03} & \textbf{\textcolor{blue}{72.03}}   & 53.58 & \textcolor{blue}{53.58}
         & \textcolor{gray}{78.46} & \textcolor{lightgray}{78.46}   & 56.81 & \textcolor{blue}{56.81} \\
         SetMask+SetPE$^{\text{Ultra}}$
         &  
         & 71.49 & \textcolor{blue}{71.49}   & \textbf{55.38} & \textbf{\textcolor{blue}{55.38}}
         & \textbf{\textcolor{gray}{80.18}} & \textbf{\textcolor{lightgray}{80.18}}   & \textbf{58.96} & \textbf{\textcolor{blue}{58.96}} \\
         
         \midrule
         Causal Mask+PE$^{\text{Ultra}}$
         &  
         & 71.55 & \textcolor{blue}{54.68}   & 53.52 & \textcolor{blue}{27.30}   
         & 64.76 & \textcolor{blue}{45.86}   & \textcolor{gray}{74.33} & \textcolor{lightgray}{63.97} \\
         \quad + Majority Vote 
         & SIQA 
         & 71.33 & \textcolor{blue}{71.33}   & 55.55 & \textcolor{blue}{55.55}
         & 65.85 & \textcolor{blue}{65.85}   & \textcolor{gray}{75.23} & \textcolor{lightgray}{75.23} \\
         SetMask+SetPE$^{\text{Ultra}}$
         &  
         & \textbf{74.16} & \textbf{\textcolor{blue}{74.16}}   & \textbf{56.83} & \textbf{\textcolor{blue}{56.83}}
         & \textbf{67.73} & \textbf{\textcolor{blue}{67.73}}   & \textbf{\textcolor{gray}{76.15}} & \textbf{\textcolor{lightgray}{76.15}} \\
         \bottomrule
         \vspace{-0.25cm} \\
         \multicolumn{10}{l}{
         $^{\text{Ultra}}$Additional pretraining
         \quad $^{\dagger}$Only first 24 (of 120) permutations tested
         }
    \end{tabular}
    \label{tab:ood}
\end{table}

\subsection{Runtimes \& memory usage}
\label{app:runtimes}

\cref{tab:runtime-memory} shows the runtimes and memory usage for Causal Mask+PE and SetMask+SetPE models with Gemma base models. All models were trained and evaluated on Nvidia H200 GPUs on an internal cluster.

The total GPU time for the paper is estimated to be around 1200 hours on Nvidia H200 GPUs.

\begin{table}[H]
    \centering
    \setlength{\tabcolsep}{3pt}
    \small
    \caption{Number of model runs per input, pre-finetuning time on Ultrafeedback, and finetuning and evaluation times on ARC-Challenge. Runtimes are calculated with Gemma 2B as the base model. $k$~is the number of (multiple) choices in the input question. Memory usage is for finetuning on ARC.
    }
    \begin{tabular}{llccccc}
         \toprule
         LLM & Model & No. Runs & Pretraining on & Finetuning on & Memory & Evaluation on \\
         & & per Input & UltraFeedback (s) & ARC (s) & Usage (GB) & ARC (s) \\
         \midrule
         \multirow{3}{*}{Gemma 2B}
         & Causal Mask+PE & $1$ & 6083.50 & 4880.68 & 13.73 & 357.63 \\
         & \quad + Majority Vote & $k!$ & - & - & - & 8620.81 \\
         & SetMask+SetPE & $1$ & 6002.45 & 4908.22 & 13.61 & 365.47 \\
         \midrule
         \multirow{3}{*}{Gemma 7B}
         & Causal Mask+PE & $1$ & 9983.51 & 7852.38 & 40.37 & 331.41 \\
         & \quad + Majority Vote & $k!$ & - & - & - & 8448.37 \\
         & SetMask+SetPE & $1$ & 10003.62 & 7821.80 & 40.02 & 329.77 \\
         \bottomrule
    \end{tabular}
    \label{tab:runtime-memory}
\end{table}

\subsection{Majority vote vs Set-LLM outputs}
\label{app:majority_vote_analysis}

\begin{figure}[ht]
    \centering
    \includegraphics[width=0.65\linewidth]{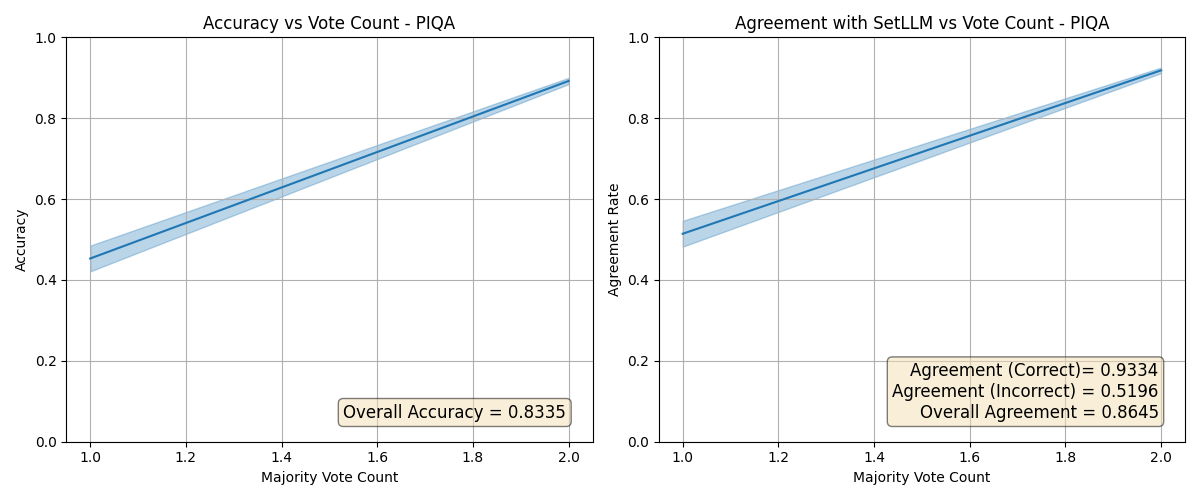}
    \caption{\emph{Causal Mask+PE$^{\text{Ultra}}$+Majority Vote} vote count versus accuracy (left). \emph{Causal Mask+PE$^{\text{Ultra}}$+Majority Vote} vote count versus agreement rate with Set-LLM (right) on PIQA.}
    \label{fig:vote-count-plots-PIQA}
\end{figure}

\begin{figure}[ht]
    \centering
    \includegraphics[width=0.65\linewidth]{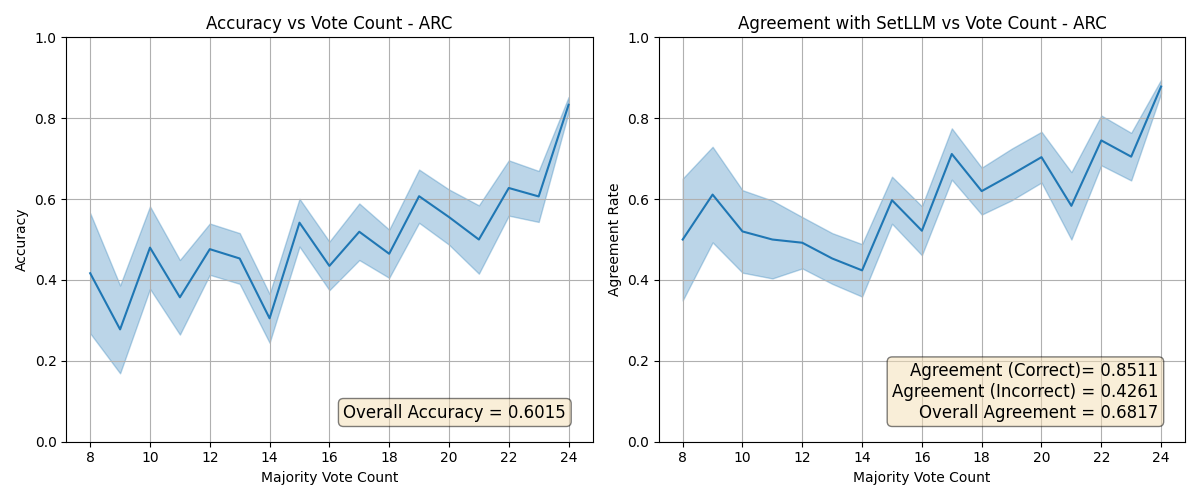}
    \caption{\emph{Causal Mask+PE$^{\text{Ultra}}$+Majority Vote} vote count versus accuracy (left). \emph{Causal Mask+PE$^{\text{Ultra}}$+Majority Vote} vote count versus agreement rate with Set-LLM (right) on ARC.}
    \label{fig:vote-count-plots-ARC}
\end{figure}

\begin{figure}[ht]
    \centering
    \includegraphics[width=0.65\linewidth]{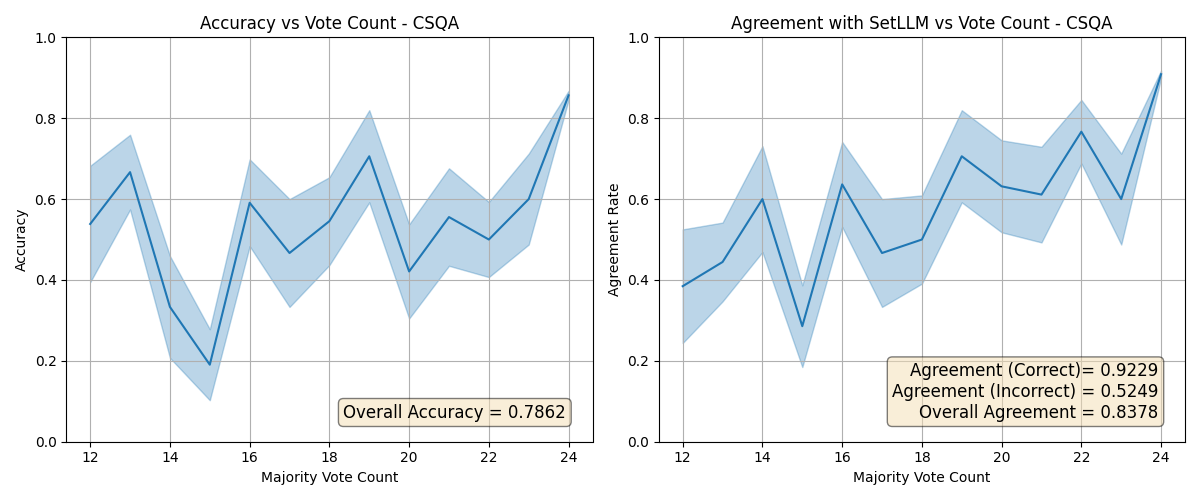}
    \caption{\emph{Causal Mask+PE$^{\text{Ultra}}$+Majority Vote} vote count versus accuracy (left). \emph{Causal Mask+PE$^{\text{Ultra}}$+Majority Vote} vote count versus agreement rate with Set-LLM (right) on CSQA.}
    \label{fig:vote-count-plots-CSQA}
\end{figure}

\begin{figure}[ht]
    \centering
    \includegraphics[width=0.65\linewidth]{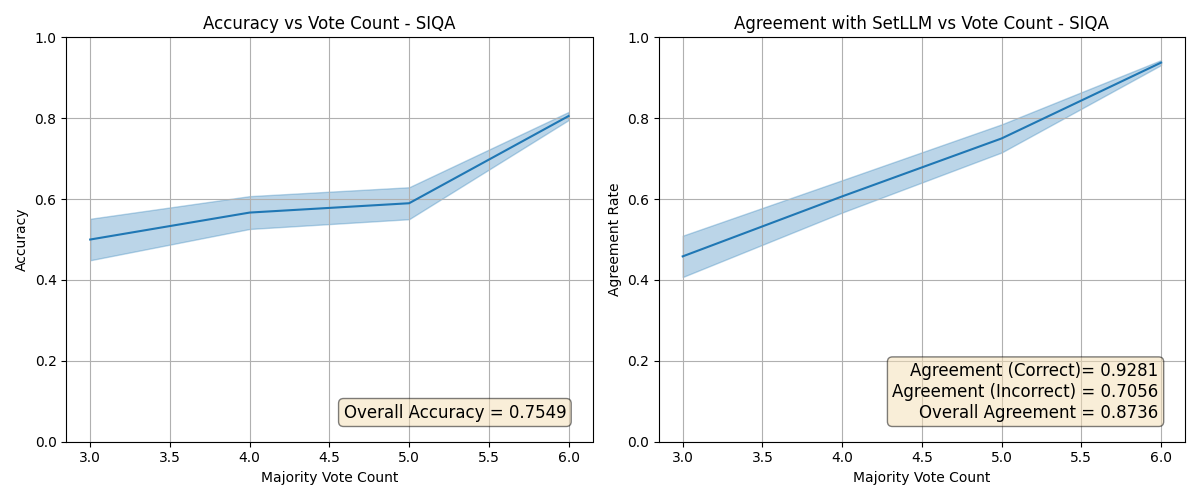}
    \caption{\emph{Causal Mask+PE$^{\text{Ultra}}$+Majority Vote} vote count versus accuracy (left). \emph{Causal Mask+PE$^{\text{Ultra}}$+Majority Vote} vote count versus agreement rate with Set-LLM (right) on SIQA.}
    \label{fig:vote-count-plots-SIQA}
\end{figure}

When calculating the majority vote, the vote count could be considered as a measure of model confidence. In \Cref{fig:vote-count-plots-PIQA,fig:vote-count-plots-ARC,fig:vote-count-plots-CSQA,fig:vote-count-plots-SIQA} (left), we compare vote count with accuracy and see that higher vote counts indeed exhibit higher accuracies on average, though the relationship is not as clear-cut for CommonsenseQA.

We explore whether our permutation-invariant model makes similar predictions to the base model, in particular for samples with high vote counts. \Cref{fig:vote-count-plots-PIQA,fig:vote-count-plots-ARC,fig:vote-count-plots-CSQA,fig:vote-count-plots-SIQA} (right) shows the agreement rate between the Gemma 2B Causal Mask+PE$^{\text{Ultra}}$ baseline and SetMask+SetPE$^{\text{Ultra}}$. We see a similar trend to the accuracy, whereby the agreement is higher for high vote count samples. However, the overall agreement rate of ($0.6817$) on ARC-Challenge demonstrates that the models often make different mistakes. Indeed the agreement rate when Causal Mask+PE$^{\text{Ultra}}$ + Majority Vote is incorrect is only around 50\% for three of the four datasets.


\end{document}